\documentclass[10pt,twocolumn,letterpaper]{article}

\usepackage[pagenumbers]{cvpr} 

\definecolor{cvprblue}{rgb}{0.21,0.49,0.74}
\usepackage[pagebackref,breaklinks,colorlinks,allcolors=cvprblue]{hyperref}
\usepackage{multirow}%
\usepackage{amsmath,amssymb,amsfonts}
\usepackage{amsthm}%
\usepackage{mathrsfs}%
\usepackage[title]{appendix}%
\usepackage{xcolor}%
\usepackage{textcomp}%
\usepackage{manyfoot}%
\usepackage{booktabs}%
\usepackage{algorithm}%
\usepackage{algorithmicx}%
\usepackage{algpseudocode}%
\usepackage{listings}%
\usepackage{graphicx}%
\usepackage{pifont}
\usepackage[utf8]{inputenc}
\usepackage{array}
\usepackage{caption}
\usepackage{pifont}
\usepackage{fontawesome} 
\usepackage{marvosym}
\usepackage{tabularx}
\usepackage{color}
\usepackage{colortbl}
\usepackage{xcolor}
\usepackage{multirow}
\usepackage{multicol}
\usepackage{wrapfig}

\newcommand{\cmark}{\ding{51}}%
\newcommand{\xmark}{\ding{55}}%
\newcommand{\rev}[1]{
    \ifthenelse{\equal{#1}{3}}{
        \faStar \faStar \faStar
    }{
        \ifthenelse{\equal{#1}{2.5}}{
            \faStar \faStar \faStarHalfO
        }{
            \ifthenelse{\equal{#1}{2}}{
                \faStar \faStar \faStarO
            }{
                \ifthenelse{\equal{#1}{1.5}}{
                    \faStar \faStarHalfO \faStarO
                }{
                    \ifthenelse{\equal{#1}{1}}{
                        \faStar \faStarO \faStarO
                    }{
                        \faStarO \faStarO \faStarO
                    }
                }
            }
        }
    }
}

\theoremstyle{thmstyleone}%
\theoremstyle{thmstyletwo}%

\theoremstyle{thmstylethree}%
\begin{document}

\title{Follow-Your-Emoji-Faster: Towards Efficient, Fine-Controllable, and Expressive Freestyle Portrait Animation}

\author{
Yue Ma$^{1*}$\quad
Zexuan Yan$^{2*}$\quad
Hongyu Liu$^{1*}$\quad
Hongfa Wang$^{3}$\quad
Heng Pan$^{3}$\quad
Yingqing He$^{1}$\quad
Junkun Yuan$^{3}$\quad \\
Ailing Zeng$^{3}$\quad 
Chengfei Cai$^{3}$\quad
Heung\mbox{-}Yeung Shum$^{1}$\quad
Zhifeng Li$^{3}$\quad 
Wei Liu$^{3}$\quad \\
Linfeng Zhang$^{2\dagger}$\quad
Qifeng Chen$^{1\dagger}$ \\
{$^{1}$HKUST}\quad
{$^{2}$SJTU}\quad
{$^{3}$Tencent} \\
$^{*}$ Equal contribution. \quad $^{\dagger}$ Corresponding author. \\
\url{https://follow-your-emoji.github.io/}
}

\vspace{-10pt}

\twocolumn[{
\maketitle

\begin{center}
    \captionsetup{type=figure}
    \includegraphics[width=\linewidth]{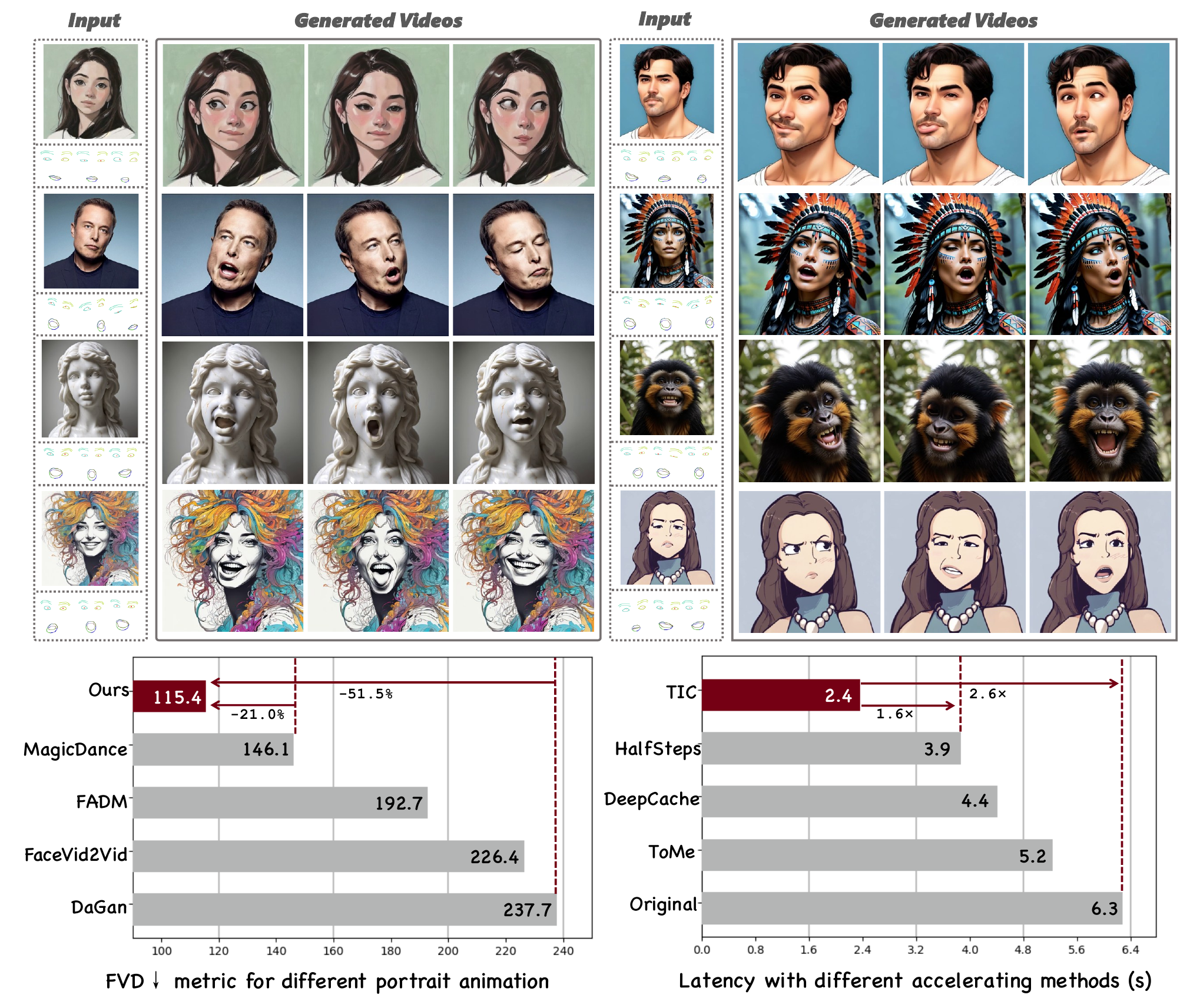}
    \captionof{figure}{\textbf{Qualitative results of our Follow-Your-Emoji-Faster.} The images of the input column are the reference portrait and the corresponding motion landmarks. 
  Using exaggerated expressions with landmark sequences, our portrait animation framework can animate freestyle reference portraits, e.g., cartoons, realism, sculptures, and even animals. Furthermore, quantitative results are shown to highlight the efficiency of our accelerating results.}
\end{center}
\label{fig:teaser}
}]

\definecolor{myred}{HTML}{bf0511}
\definecolor{myblue}{HTML}{1272b6}
\definecolor{myyellow}{HTML}{000000}

\begin{abstract}
We present Follow-Your-Emoji-Faster, an efficient diffusion-based framework for freestyle portrait animation driven by facial landmarks. The main challenges in this task are preserving the identity of the reference portrait, accurately transferring target expressions, and maintaining long-term temporal consistency while ensuring generation efficiency. To address identity preservation and accurate expression retargeting, we enhance Stable Diffusion with two key components: a expression-aware landmarks as explicit motion signals, which improve motion alignment, support exaggerated expressions, and reduce identity leakage; and a fine-grained facial loss that leverages both expression and facial masks to better capture subtle expressions and faithfully preserve the reference appearance. With these components, our model supports controllable and expressive animation across diverse portrait types, including real faces, cartoons, sculptures, and animals. However, diffusion-based frameworks typically struggle to efficiently generate long-term stable animation results, which remains a core challenge in this task. To address this, we propose a progressive generation strategy for stable long-term animation, and introduce a Taylor-interpolated cache, achieving a \textbf{2.6×} lossless acceleration. These two strategies ensure that our method produces high-quality results efficiently, making it user-friendly and accessible. Finally, we introduce EmojiBench++, a more comprehensive benchmark comprising diverse portraits, driving videos, and landmark sequences. Extensive evaluations on EmojiBench++ demonstrate that Follow-Your-Emoji-Faster achieves superior performance in both animation quality and controllability.
The code, training dataset and benchmark will be found in \url{https://follow-your-emoji.github.io/}.
\end{abstract}

\section{Introduction}

We address the problem of portrait animation, in which the pose and expression dynamics extracted from a driving video are transferred onto a static reference portrait. Recent advances employing GANs~\cite{goodfellow2020generative} and diffusion models~\cite{sohl2015deep} have shown remarkable capabilities across diverse applications such as interactive video conferencing, virtual avatar creation, and augmented reality.

For GAN-based portrait animation methods~\cite{megaPortriat, wang2021facevid2vid, fomm, liu2023human}, a two-stage framework is commonly employed. In the first stage, a learned flow field warps the reference image in feature space; in the second stage, a GAN-based rendering decoder refines these warped features and generates any missing or occluded regions. However, the intrinsic limitations of GANs and inaccuracies in flow-based motion representation often lead to results beset by unrealistic details and conspicuous artifacts.

More recently, diffusion models~\cite{ho2020denoising, song2020denoising} have demonstrated superior generative performance over GANs. Several approaches leverage large-scale foundation diffusion models for high-fidelity video synthesis~\cite{guo2023animatediff, blattmann2023align, wang2024cove, feng2025dit4edit, ho2022imagen, ho2022video, wu2023tune, he2022lvdm, chen2024videocrafter2, chen2023videocrafter1} and photorealistic image generation~\cite{rombach2021highresolution, saharia2022photorealistic, ramesh2022hierarchical}. Nevertheless, these pretrained models are not specifically tailored to portrait animation, as they struggle to preserve the reference portrait’s identity and to accurately transfer nuanced expressions.

Several recent works~\cite{magicpose, xu2023magicanimate, zhu2024champ, hu2023animate, wang2023disco} extend large-scale diffusion frameworks (e.g., Stable Diffusion~\cite{rombach2021highresolution}) by inserting plug‑and‑play modules tailored for portrait animation. These methods typically employ an appearance network~\cite{hu2023animate} together with CLIP~\cite{radford2021learning} to encode identity characteristics from the reference portrait, and leverage temporal attention to ensure inter‑frame coherence. Despite these adaptations, the generated videos often display distortions and unrealistic artifacts when animating out‑of‑distribution subjects (e.g., cartoons, sculptures, animals). We pinpoint two core issues: (1) The motion representations—whether 2D landmarks~\cite{magicpose, hu2023animate} or motion maps~\cite{xie2024x}—lack sufficient robustness. Landmark‑based guidance can introduce spatial misalignments that lead to identity drift, while motion maps (as in Xportrait~\cite{xie2024x}) require external identity swaps during training, which disrupts subtle expression details. (2) These pipelines optimize the standard diffusion loss, which does not explicitly enforce precise modeling of facial appearance and expression dynamics. Furthermore, diffusion‑based architectures suffer from low inference efficiency, hindering real‑time interaction, and often fail to maintain temporal stability over long sequences, limiting their practical usability.

In this work, we introduce \textit{Follow-Your-Emoji-Faster}, a diffusion‑based framework for portrait animation. Building on the appearance network and temporal attention modules common to recent diffusion approaches, we incorporate several novel components to overcome the limitations discussed above. 
(1) \textbf{Expression‑Aware Landmarks.} We propose a new control signal obtained by projecting 3D facial keypoints from MediaPipe~\cite{mediapipe} into the image plane. Thanks to the canonical nature of these 3D points, our landmarks remain consistently aligned with the reference portrait during inference, effectively preventing identity drift. To mitigate occasional inaccuracies in MediaPipe’s facial contour estimation, we exclude contour points and include pupil landmarks only, ensuring the network focuses on expression dynamics (e.g., pupil motion) without distorting the underlying identity.
(2) \textbf{Facial Fine‑Grained Loss.} To encourage accurate capture of subtle expressions and detailed appearance, we define a loss over both facial and expression masks derived from our expression‑aware landmarks. Specifically, we compute the per‑pixel spatial distance between prediction and ground truth within these mask regions, guiding the model to prioritize fine expression changes.
These enhancements enable high‑quality, free‑style portrait animation (see Fig.~\ref{fig:teaser}). To scale beyond single‑step diffusion limitations for long videos—where GPU memory can be prohibitive—we adopt a progressive generation strategy, yielding stable, high‑fidelity long‑term animations.

Finally, although our model achieves satisfying generation quality, there is still room for improvement in inference speed. In recent years, various acceleration techniques for diffusion models have emerged and demonstrated impressive speedups in image and video generation tasks. However, these methods have not been thoroughly explored or applied in conditional editing tasks involving additional pose and signal guidance.
In our framework, existing acceleration approaches fail to address two key challenges: (1) the preservation of region-specific key information guided by landmarks, and (2) the varying rates of feature change across different denoising stages.
To address these issues, we propose a novel caching-based acceleration strategy for our pipeline, Taylor-Interpolated Cache (TIC). Specifically, TIC leverages the spatial distribution knowledge provided by landmarks and adopts a finer-grained update strategy on corresponding positional masks. The cached features are not only reused to accelerate inference but are also refined at later timesteps using a Taylor expansion-based interpolation strategy.
Compared to other acceleration methods, TIC achieves the best lossless quality. It delivers a notable speedup of over 2.6×, enabling our approach to serve a broader range of practical applications.

To train our framework, we assemble a high‑quality expression dataset comprising 18 exaggerated facial expressions and 20‑minute real‑human videos from 115 participants. We also leverage a progressive generation strategy to support long‑duration animation synthesis with both stability and high fidelity. For evaluation, we introduce EmojiBench++, an extension of the original EmojiBench~\cite{Followyouremoji}, containing \textcolor{myyellow}{500} diverse portrait animation sequences that span a broad spectrum of expressions and head orientations. We further benchmark Follow-Your-Emoji-Faster on EmojiBench++ and compare it against existing methods. Our experiments show that our approach outperforms prior baselines in quantitative metrics and qualitative assessments, delivering superior visual quality, faithful identity preservation, precise motion transfer and efficient generation latency, even for out‑of‑domain portraits and movements.

In summary, our contributions can be summarized as follows:

\begin{itemize}

\item We introduce \textit{Follow-Your-Emoji-Faster}, a diffusion-based framework for fine-controllable portrait animation. Based on the proposed progressive generation strategy, it can further produce long-term animation effectively.

\item 
To facilitate freestyle portrait animation, we propose the expression-aware landmarks as the motion representation and a facial \textcolor{black}{fine-grained} loss to help the diffusion model enhance the generation quality of facial expressions.

\item To train our model, we introduce a new expression training dataset with 18 expressions and 20-min talking videos from 115 subjects. To validate the effectiveness of our methods, we construct an improved benchmark EmojiBench++. Comprehensive results show the superiority of our Follow-Your-Emoji-Faster in fine-controllable, expressive aspects.

\item \textcolor{black}{To accelerate the inference speed of our model, we introduce an innovative interpolation-based caching strategy, Taylor-Interpolated Cache (TIC). TIC fully leverages both temporal and spatial information, achieving a lossless speedup of over 2.6×.}

\end{itemize}

\section{Related Work}

\subsection{Single-forward Portrait Animation}
The goal of portrait animation is to bring a static portrait image to life by leveraging given driven signals, including a sequence of facial landmarks or portrait video frames. It has attracted a lot of attention in the research. Previous approaches~\cite{megaPortriat, ma2025controllable, fomm, magicpose, averbuch2017bringing, thies2016face2face, kim2018deep, wiles2018x2face} mainly leverage Generative Adversarial Networks (GANs)~\cite{goodfellow2020generative} to generate plausible motion using self-supervised learning.  
Due all GAN method methods can work without a discriminator, we consider them as the single-forward methods. 
The pioneering works primarily involved two steps: warping and rendering. These methods firstly estimate head and facial motion with open-source 2D/3D pose predictors~\cite{mediapipe, dwpose}. The facial representation is warped and fed into a generative model to synthesize dynamic frames with realistic animation and rich details.
Following such a paradigm, a majority of approaches~\cite{wang2021facevid2vid, hong2022depth, ma2025followcreation, TPS, ReenacArtFace} focus on improving facial warping estimation, including 3D neural landmarks~\cite{wang2021facevid2vid}, thin-plate splines~\cite{TPS} and depth~\cite{hong2022depth}. 
Additionally, the 3D morphable is utilized to model the expression and motion in ReenacArtFace~\cite{ReenacArtFace}. ToonTalker~\cite{gong2023toontalker} employs the transformer architecture to help the warping process of cross-domain datasets. 

MegaPortraits~\cite{megaPortriat} enhances rendered image quality using high-resolution image data, whereas FADM~\cite{zeng2023face} enriches generated details using the proposed coarse-to-fine animation framework. Face Vid2Vid~\cite{wang2021facevid2vid} presents a pure neural rendering to decompose identity-specific and motion-related information unsupervisedly.
In addition to video reenactment, there are also various driving signals, such as 3D facial prior~\cite{deng2020disentangled, fried2019text, feng2021learning, ma2025followyourmotion, khakhulin2022realistic, sun2023next3d, xu2023omniavatar} and audio~\cite{tian2024emo, xu2024vasa, he2023gaia, zhang2023sadtalker}. However, these methods primarily focus on talking scenarios, and they struggle to synthesize animated frames with high-quality facial details. Meanwhile, they often meet the significant challenge when there is a large style domain gap between source portrait image and training data, limiting the practical application in diverse domain styles. In contract, our approach enable to animate the freestyle portrait using given signals, including cartoons, realism, sculptures, and even animals.

\subsection{Diffusion-based Portrait Animation}

With the development of diffusion models (DMs)~\cite{ho2020denoising, song2020denoising},
current approaches achieve superior performance in various generative tasks including image generation~\cite{rombach2021highresolution, zhao2019image, xiao2023semanticac, ruiz2023dreambooth} and editing~\cite{cao2023masactrl, liu2025avatarartist, long2025follow, hertz2022prompt, xiao2024bridging, brooks2023instructpix2pix}, video generation~\cite{ma2024follow, feng2025follow, zhang2025magiccolor, singer2022make, he2022latent, zhu2024instantswap, ma2022visual, ma2024followyourpose, ma2024followyourclick} and editing~\cite{qi2023fatezero, zhang2023controlvideo, liu2023video, ma2023magicstick}. Recently, latent diffusion models further improved the performance by operating the diffusion step in latent space. 
Mainstream portrait animation approaches leverage the power of Stable Diffusion (SD)~\cite{rombach2021highresolution} and incorporate temporal information into generation process, such as AnimateDiff~\cite{guo2023animatediff}, MagicVideo~\cite{zhou2022magicvideo}, VideoCrafter~\cite{chen2023videocrafter1} and ModelScope~\cite{wang2023modelscope}.
Additionally, to preserve appearance context in the original image,  many works try to ~\cite{magicpose, xu2023magicanimate, hu2023animate, zhu2024champ} inject the reference image into the self-attention blocks in the LDM UNets, facilitating image editing and video generation. 
Animateanyone~\cite{hu2023animate} proposes the dual-UNet structure to inject the original portrait, preserving the identify consistency.  AvatarArtist~\cite{liu2025avatarartist} propose a diffusion  based method for open domain 4D avatar generation.

\textcolor{myyellow}{While recent models achieve impressive video quality, they rely predominantly on text prompts for semantic guidance, which can be vague and fail to capture precise user intent. To remedy this, a variety of control signals—such as structural cues~\cite{gen2,xing2024make}, pose information~\cite{magicpose,ma2024follow}, and Canny edges~\cite{zhang2023controlvideo}—have been incorporated for controllable video synthesis. Notably, several concurrent approaches~\cite{magicpose,xu2023magicanimate,zhu2024champ,hu2023animate} attain state-of-the-art full-body animation by seamlessly embedding appearance and motion controls within temporal attention modules. However, these frameworks predominantly target full-body motion and overlook fine-grained facial details. In contrast, we present a diffusion-based architecture tailored to animate diverse portrait styles with rich facial fidelity (e.g., eye movements and skin textures).}

\subsection{Acceleration of Diffusion Models}
Recent advancements in diffusion models have brought more opportunities to filmmaking and advertising, meeting the rapid development needs of the various industry. 
However, the long inference time of diffusion models is a key challenge, which restricts their practical applications.
Currently,  the mainstream diffusion model acceleration paths are divide to two type: sampling step reducing and internal computation acceleration. As for sampling step reduction~\cite{Song2022DDIM,lu2022dpmsolverfastodesolver,lu2023dpmsolverfastsolverguided,liu2022flowstraightfastlearning}, 
\textcolor{black}{
Through mathematical analysis of the sampling path, these methods significantly reduce the number of sampling steps while preserving sampling quality, thereby accelerating inference.
}
Consistency models~\cite{song2023consistency} accelerate diffusion model generation by learning the consistency of data distributions across different noise levels in the latent space. For internal computation acceleration, current approaches mainly include network compression~\cite{Xie2024SANA}, token pruning~\cite{zhang2025sito}, token merging~\cite{bolya2023tokenmergingvitfaster,Feng2024DiT4Edit, bolya2023tomesd}, and layer-wise caching techniques~\cite{deepcache,gao2024dual,L2C,Liu2024FasterDiffusion,fora,liu2025taylorseers}.
Nevertheless, layer-wise caching techniques exhibit a large cache granularity. These approaches overlook the asymmetry of importance at the token level~\cite{yan2025eeditrethinkingspatial}. ToCa~\cite{toca} and DuCa~\cite{duca} employ a token-wise cache approach. They focus on tokens and assign importance using score maps. During recomputation processing, a specific proportion of tokens are chosen for refreshing, thereby achieving more precise control over caching process.
\textcolor{black}{
However, existing acceleration methods are not well-suited for face pose-controlled video generation tasks. They fail to fully exploit both the temporal relationships and differences in model sampling results across timesteps, as well as the additional spatial information provided by facial landmarks.
In contrast, we adopt a dynamically adjusted cache update mechanism across timesteps, while fully leveraging facial landmark information to more precisely control cache updates in important regions.
}

\section{Preliminaries}

\subsection{Latent Diffusion Model} 
\textcolor{myyellow}{Latent Diffusion Models (LDMs)~\cite{rombach2021highresolution}, the core architecture behind Stable Diffusion, perform diffusion and denoising in a lower-dimensional latent space rather than directly in pixel space, enabling more stable and efficient training. Input images are first encoded into latent representations by a VAE~\cite{kingma2013auto}, and these embeddings are then conditioned on textual prompts during the diffusion process.}
A U-Net backbone~\cite{ronneberger2015u}, augmented with Transformer blocks for both self‑attention and cross‑attention, learns the reverse denoising process in the latent space. The cross‑attention layers allow the textual prompt to be injected effectively at each step. The U-Net’s training objective is defined as:
\begin{equation}
\mathcal{L}_{LDM}=\underset{t, z_, \epsilon}{\mathbb{E}} [ \|\epsilon-\epsilon_\theta (\sqrt{\bar{\alpha}_t} z +\sqrt{1-\bar{\alpha}_t} \epsilon, c, t ) \|^2 ],
\label{eq:ldm}
\end{equation}
where $\mathbf{z}$ denotes the latent representation of a training sample $\mathbf{x}$. The functions $\epsilon_\theta$ and $\epsilon$ correspond to the noise predicted by the U‑Net and the actual Gaussian noise added at timestep $t$, respectively. The vector $\mathbf{c}$ contains the embeddings of all conditioning inputs. Finally, $\bar{\alpha}_t$ is the same scheduling coefficient used in standard diffusion models.

\subsection{Portrait Animation with Diffusion}

\textcolor{myyellow}{Recent methods \cite{magicpose, xu2023magicanimate, zhu2024champ, hu2023animate, xie2024x,echomimic,xu2025hunyuanportrait,cui2024hallo2,loopy} adopt a shared, modular architecture to adapt pre-trained Stable Diffusion (SD) for full-body and portrait animation. Their frameworks typically consist of four plug-and-play components:
\textbf{1. Prompt Injection:} An image encoder replaces the CLIP text encoder to extract tokens from the reference portrait. These tokens are then fed into the UNet via cross-attention layers, analogous to text prompts in SD \cite{xu2023magicanimate, liu2023stylecrafter, hu2023animate}.
\textbf{2. Appearance Network:} Structurally identical to SD’s UNet, this module captures identity-specific attributes and background context from the reference image, injecting its features into self-attention blocks to preserve appearance \cite{magicpose}.
\textbf{3. Temporal Attention:} Temporal transformer layers are integrated into the UNet to model cross-frame correspondences and ensure coherent animation across time.
\textbf{4. Motion Control Injection:} Spatial control signals—such as pose maps or motion vectors—are incorporated either via ControlNet \cite{zhang2023adding} or by directly appending motion features to the UNet inputs \cite{hu2023animate}.}

\section{Method}
\label{sec:Method}
\begin{figure*}
  \centering
  \includegraphics[width=\textwidth]{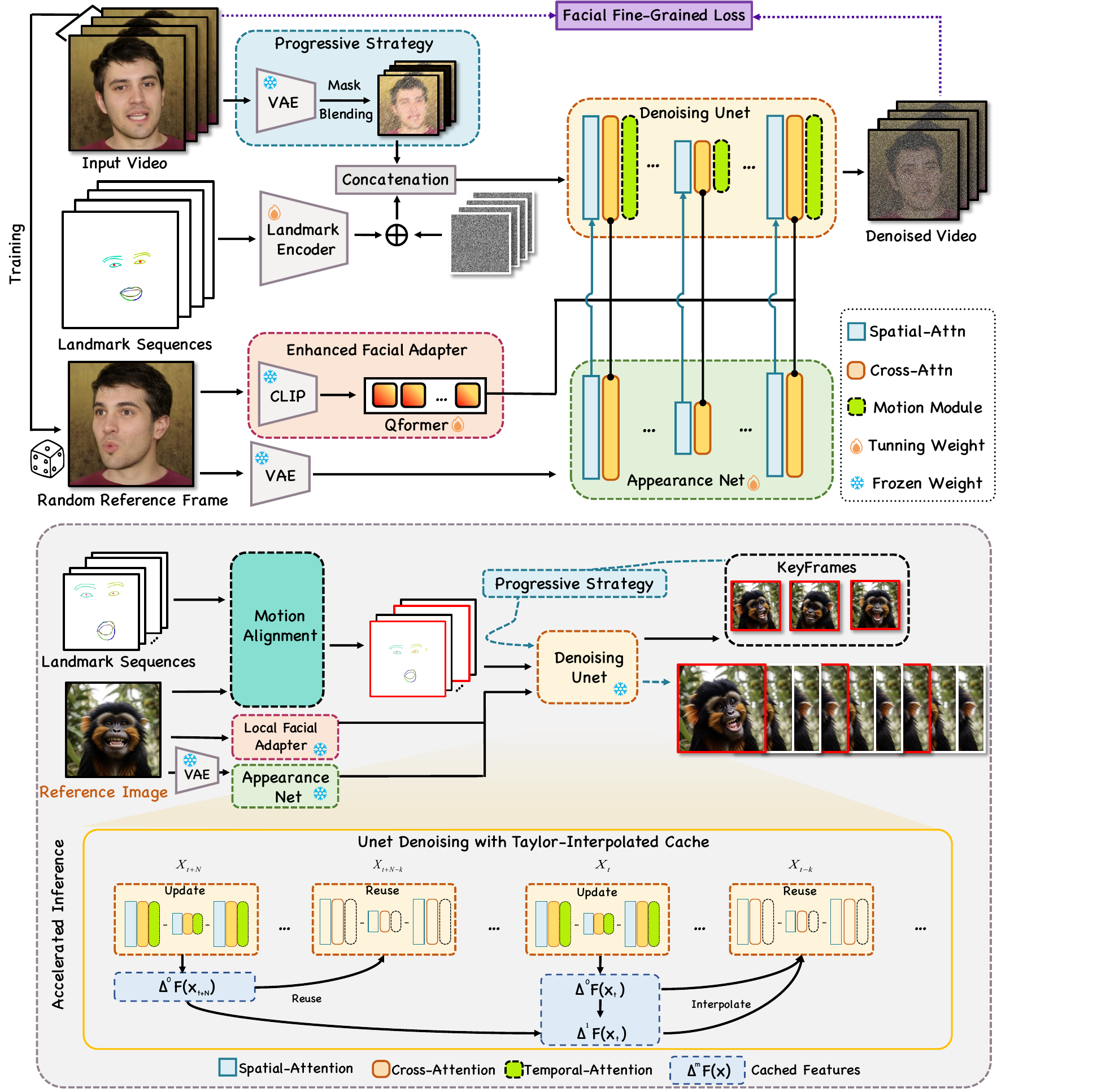} 
  \caption{\textbf{The overview of Follow-Your-Emoji-Faster}. We extract the features of our expression-aware landmark sequence with a landmark encoder and fuse these features with multi-frame noise first, then we utilize the progressive strategy to mask the frame of the input latent sequence randomly.  Finally, we concatenate this latent sequence with the fused multi-frame noise and feed it to the Denoising UNet to conduct the denoising process for video generation. The appearance net and image prompt injection module help our model preserve the identity of the reference portrait, and the temporal attention maintains the temporal consistency. During training, the facial fine-grinded loss guides the Unet to pay more attention to the facial and expression generation. During inference, we align the target landmark with the reference portrait with the motion alignment module. \textcolor{myyellow}{Then, we  generate the keyframes and utilize the progressive strategy to predict long videos with Taylor-Interpolated Cache, which accelerate inference process via reusing and predicting layer-wise features.} }
  \label{fig:framework}
\end{figure*}

\textcolor{myyellow}{
The overall pipeline of our approach is illustrated in Fig.~\ref{fig:framework}. First, from an input video clip we randomly sample one frame, denoted as the reference portrait $\mathcal{I}_0$, and extract an expression-aware landmark sequence $\{L_1, L_2, \dots, L_N\}$ from the remaining frames. To ensure compatibility with diverse portrait styles—including cartoons, sculptures, and animals—we apply a motion alignment step: 3D landmarks are detected using MediaPipe \cite{mediapipe}, a projection matrix is computed between the driving and target landmarks, and this retargeting matrix is used to warp each landmark. Eye motion is handled by computing the relative position of the eye centers and applying the same transformation to the pupil coordinates.
Next, we follow recent diffusion-based portrait animation frameworks by employing an appearance network and temporal attention within the Stable Diffusion UNet. Control signals are injected by encoding our expression-aware landmarks with a landmark encoder and directly adding these features into the UNet input. In parallel, the reference image $\mathcal{I}_0$ is passed through a pre-trained CLIP image encoder to obtain image tokens, which are then fused via a four-layer Q-Former \cite{zhang2023vision}—a design similar to prior works \cite{liu2023stylecrafter, ye2023ipadapter, wang2024instantstyle, wang2024instantstyle1}.}
\textcolor{myyellow}{
Additionally, we introduce a training-free inference framework that accelerates pre-trained models via layer-wise feature caching. Given a sequence of diffusion timesteps \(\{T_0, T_{1}, \dots, T_\tau\}\), we store the fully computed feature maps every \(\mathcal{N}\) steps. To reduce drift and maintain high video quality, we update these cached features after \(\mathcal{G}\) steps using a Taylor-based numerical interpolation scheme.}

In the following sections, we first define our motion representation and detail the construction of expression-aware landmarks in Sec.~\ref{sec:Expression Aware Landmark}. We then introduce our facial fine-grained loss in Sec.~\ref{sec:Facial Fine-grained loss}. To address challenges in long-term animation, we describe our progressive synthesis strategy in Sec.~\ref{sec:Progressive Strategy}. \textcolor{myyellow}{Finally, we propose a training-free caching technique in Sec.~\ref{sec:TIC}.}

\subsection{Expression-Aware Landmark} 
\label{sec:Expression Aware Landmark}

\textcolor{myyellow}{
Motion representation of facial expressions plays a pivotal role in portrait animation, as it governs the fidelity with which subtle emotional cues are conveyed. Existing diffusion-based approaches typically employ either sequential portrait frames as the driving signal or 2D facial landmarks extracted from these frames~\cite{xie2024x}. However, relying on 2D landmarks at inference can lead to misregistration between the intended expression and the static reference portrait, degrading accuracy and risking identity leakage. Conversely, using full image sequences as motion inputs circumvents misalignment but mandates that, during training, the driving subject differ from the reference identity—necessitating a preliminary identity-conversion step via a secondary animation model. This conversion not only compromises expression fidelity but also fails when the source and target belong to disparate categories (e.g., animating a canine visage with human features).}
\textcolor{myyellow}{
To overcome these limitations, we propose an expression-aware landmark representation derived from 3D keypoints. First, we employ MediaPipe to extract dense 3D facial landmarks from a motion video, then orthogonally project them onto the image plane, discarding contour points while preserving only those corresponding to salient facial features (see Fig.~\ref{fig:ablation_lmk}). This selective projection sharpens the model’s focus on nuanced deformations and mitigates errors arising from large contour displacements. Additionally, we compute the relative positions of the irises within the 3D eye sockets and maintain this relationship post-projection to capture fine-grained gaze motion. Because our landmarks originate in MediaPipe’s canonical 3D space, we can naturally align any target landmark sequence to the static reference portrait during inference (motion alignment), as illustrated in Fig.~\ref{fig:framework}.}

\begin{figure}
  \centering
  \includegraphics[width=\linewidth]{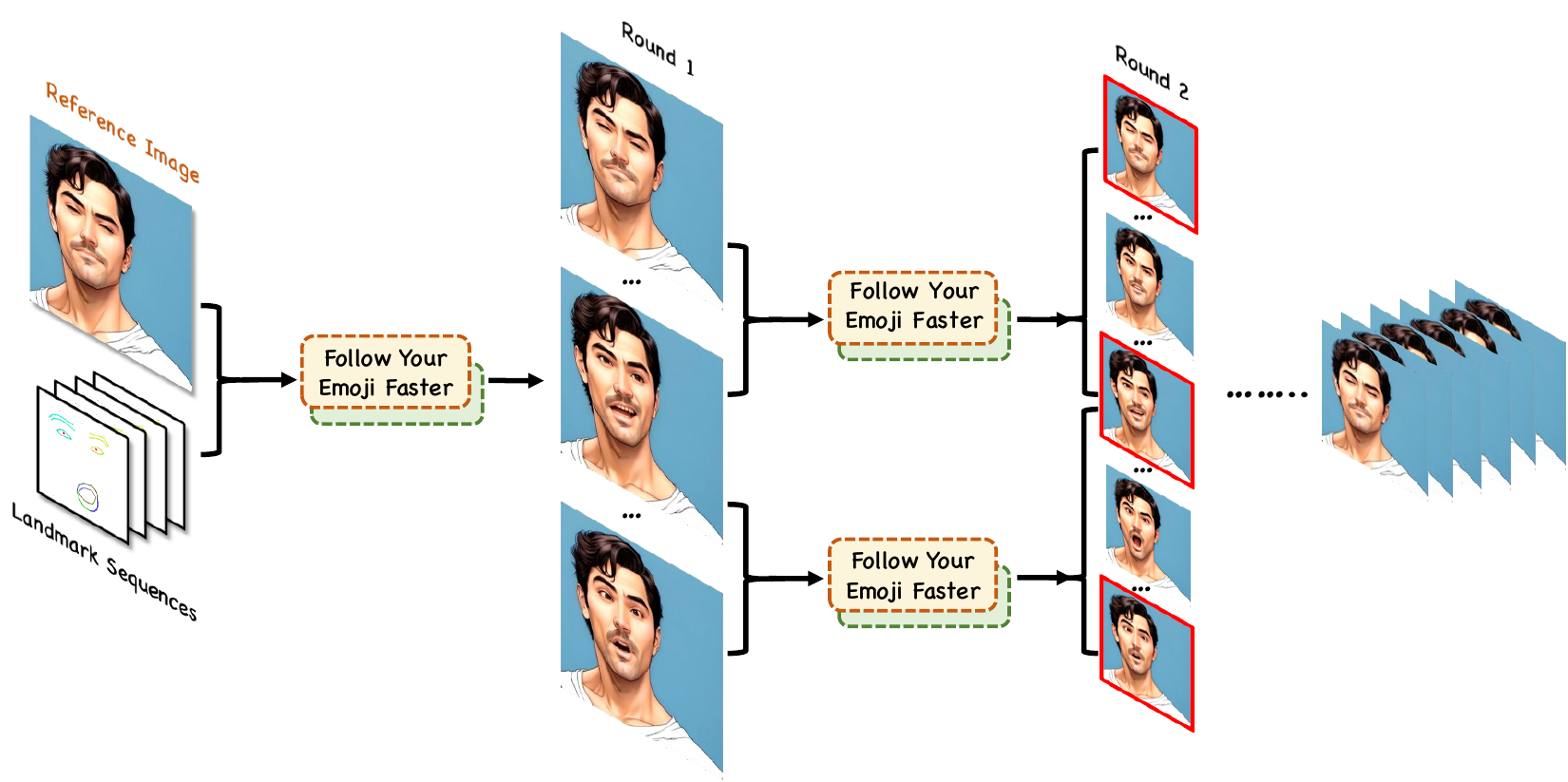} 
  \caption{\textcolor{black}{\textbf{The illustration of progressive strategy.} Similar to the training stage, we first generate the keyframes of the video, then we concatenate the first and last frames to the noise and input them into the model to generate the intermediate content.}}
  \label{fig:progress}
\end{figure}

\subsection{ Facial Fine-Grained Loss}
\label{sec:Facial Fine-grained loss}
\begin{figure}[t]
  \centering
  \includegraphics[width=\linewidth]{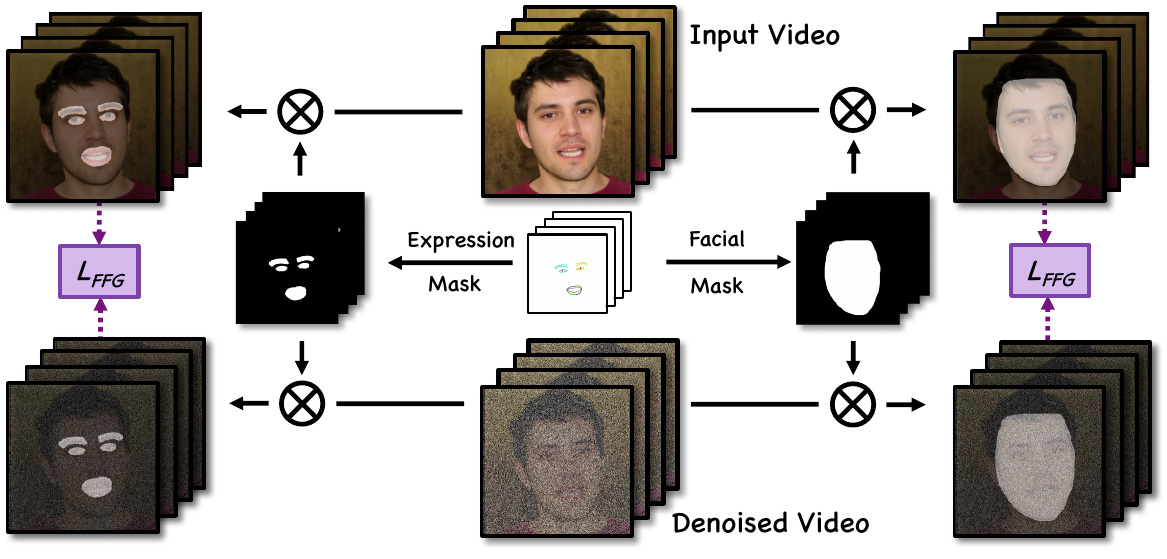} 
  \caption{\textbf{The detail of our facial fine-grained loss.} We extract the facial mask and expression mask with our landmark first. Then, we calculate the denoising loss $\mathcal{L}_{FFG} $  in these masked regions.  }
  \vspace{-0.18in}
    \label{fig:ffgloss}
\end{figure}
For the portrait animation task, we hope the diffusion model focuses on expression generation and identity preservation. However, the diffusion model's original training objective $\mathcal{L}_{LDM}$ is to learn the content of all regions of the target image, which has no specific constraints for learning the facial content during the training process. Therefore,  we propose the facial fine-grained (FFG) loss to modify the $\mathcal{L}_{LDM} $ and make the model pay more attention to the content of facial and expression regions.

As shown in Fig.~\ref{fig:ffgloss}, we need to get two types of masks to capture the expression and facial regions to calculate the FFG loss.  For the expression mask ${\mathcal{M}_e} $,  we dilate each point of our expression-aware landmark and set these dilation regions as the expression mask. For the facial mask ${\mathcal{M}_f} $, we project the MediaPipe 3D facial contour's keypoints and connect these projected points to get the facial masks. Finally, these two masks split the FFG loss into expression and facial aspects, respectively. 
Formally, the loss function can be written as below:
\begin{equation}
\begin{aligned}
\mathcal{L}_{FFG}={\mathbb{E}}\left[\left\|\mathcal{M}_e \cdot \left(z-\hat{z} \right)+\mathcal{M}_f \cdot \left(z-\hat{z}\right)\right\|^2\right],
\label{eq:ffg_loss}
\end{aligned}
\end{equation} where $\hat{z}$ is the prediction latent embedding obtained by decoding the $\epsilon_{\theta}$. With our FFG loss, our method demonstrates better performance in both identity preservation and expression generation, as shown in Fig.~\ref{fig:ablation_ffgloss}. Finally, our total loss can be written as
\begin{equation}
    \begin{aligned}
        \mathcal{L} =\mathcal{L}_{LDM}+\mathcal{L}_{FFG}.
    \end{aligned}
\end{equation}

\subsection{Progressive Strategy for Long-Term Animation}
\label{sec:Progressive Strategy}

\textcolor{myyellow}{
With the rapid evolution of animation technologies and growing user expectations, producing temporally coherent long-duration videos has become critical. Although prior methods~\cite{magicpose, xu2023magicanimate, zhu2024champ, hu2023animate} train on short clips, they extend to longer sequences at test time by stitching overlapping segments and applying Gaussian smoothing—an approach we find compromises temporal consistency.}

\textcolor{myyellow}{
To address this, we design a coarse‑to‑fine progressive generation strategy for long‑term animation. Fig.~\ref{fig:progress} details our pipeline. During inference, we first treat the initial and terminal frames as keyframes, then synthesize intermediate frames via interpolation guided by these anchors. To emulate this at training time, we mask all latent representations except the first and last frames, concatenate the masked sequence with the UNet’s inputs, and perform the denoising step; this encourages the network to reconstruct content for keyframes given sparse context. Simultaneously, we apply random masking to each latent frame with a 50\% probability, ensuring that every frame receives reconstruction supervision across training iterations. These two masking schemes are alternated with equal probability throughout training, thereby preparing the model for efficient, high‑fidelity long‑term animation generation.
}

\subsection{Efficient Inference via Feature Caching}
\label{sec:TIC}

\begin{algorithm}[ht]
\caption{\textbf{Accelerated Unet Inference}}
\label{alg:unet_accelerate}
\begin{algorithmic}[1]
\Require 
Ref-features \(R_f\), Input latents \(\mathbf{Z_0}\)
Landmark regions $\mathcal{M}_{lmk}$, Hyperparameter $\mathcal{G}$,
Cache \(\mathcal{C}[tag,t]\), and Refresh interval N.

\Ensure 
Output latents \(\mathbf{Z_T^*}\).

\State // \textbf{Denoising Process with Caching Features}
\For{\(t=0,1\dots T\)}
    \For{$\mathcal{F} \in$  \texttt{UNet-Blocks}}
        \State // \textbf{Feature Access Rule}
        \If{t \% N = 0}
            \State $\tilde{\mathbf{Z}}_{t} \gets \mathcal{F}(\mathbf{Z}_t,R_f) ,k\gets0$
        \Else
            \State \(\tilde{\mathbf{Z}}_{t} \gets \mathcal{C}[\mathcal{F},t-1], k\gets k+1\)
        \EndIf
        \State // \textbf{Feature Update Rule (Eq. \ref{eq:4})}
        \If{$t \geq \mathcal{G}$ \textbf{or} $z_t \in \mathcal{M}_{lmk}$}
            \State $\displaystyle \mathcal{C}[\mathcal{F},t] \gets \tilde{\mathbf{Z}}_t + \sum_{i=1}^\mathcal{O} \frac{\Delta^i\mathcal{F}(\mathbf{Z}_t,R_f)}{i! \cdot N^i}k^i$
        \Else
            \State $\mathcal{C}[\mathcal{F},t] \gets \tilde{\mathbf{Z}}_t$
        \EndIf
    \EndFor
\EndFor

\State \(\mathbf{Z}_T^* \gets \texttt{Post-Process}(\mathbf{Z_T})\)
\State \Return $\mathbf{Z}_T^*$
\end{algorithmic}
\end{algorithm}

\textcolor{black}{
 \textcolor{myyellow}{By storing intermediate feature maps, we alleviate the U-Net's computational burden during denoising, thereby enabling efficient generation (see bottom of Fig.~\ref{fig:framework}).}
 We design a corresponding caching algorithm~\ref{alg:unet_accelerate} that fully leverages both historical cache information and the spatial structure provided by the facial landmark mask. Unlike traditional caching schemes that reuse features statically, our TIC dynamically employs Taylor expansion–based interpolation to update and access more precise feature representations from the cache. A hyperparameter \( \mathcal{G} \) is introduced to specify the timestep boundary where the Taylor interpolation mode is applied. It is set to be slightly above 70\% of the inference time steps, to indicate the ending stage in the denoising process. The mask \( \mathcal{M}_{lmk} \) corresponds to facial landmarks in the feature map, and is extracted in a manner consistent with the approach illustrated in Fig.~\ref{fig:ffgloss}, while the background region is expressed as \(\mathbb{U}/\mathcal{M}_{lmk}\). The core update rule for TIC can be formalized as follows:}
\vspace{-2em}
\begin{equation}
\resizebox{\linewidth}{!}{$
\mathcal{F}(z^{}_{t+k}) \gets 
\begin{cases}
\mathcal{F}(z^{}_{t}), \qquad \qquad \qquad t<\mathcal{G} \text{ and }z\in \mathbb{U}/\mathcal{M}_{lmk}  \\
\mathcal{F}(z^{}_{t}) + \displaystyle \sum_{i=1}^{\mathcal{O}} \frac{\Delta^{i}\mathcal{F}(z^{}_{t})}{i! \cdot N^{i}} k^i, \quad t \ge \mathcal{G} \text{ or }z\in \mathcal{M}_{lmk}.
\end{cases}
$}
\label{eq:4}
\end{equation}
\textcolor{black}{
In Equation~\ref{eq:4}, the hyperparameter \(\mathcal{G}\) is employed to differentiate the threshold between the early and final stages of the denoising process.
\( \mathcal{F} \) represents modules within downsampling and upsampling blocks, including Spatial-Attention, Cross-Attention and Temporal-Attention. The difference operator \( \Delta \) is defined recursively to capture temporal changes in feature representations.
}
\textcolor{black}{
The first-order difference is defined as
\(
\Delta \mathcal{F}(z_t) = \mathcal{F}(z_t) - \mathcal{F}(z_{t - N})
\), and the \( n \)-th order recursive difference is defined as
\(
\Delta^n \mathcal{F}(z_t) = \Delta^{n-1} \mathcal{F}(z_t) - \Delta^{n-1} \mathcal{F}(z_{t - N})
\), 
where \( N \) denotes the stride of cached timesteps.
}

\textcolor{black}{
The cache design takes into account the observation that, during the denoising process, feature dynamics are more volatile in the early stages~\cite{delta_dit}. In this phase, predictions based on Taylor expansion are more prone to failure due to rapidly changing representations. In contrast, in the ending stage of denoising, the model's updates become more stable and gradual, allowing Taylor-based interpolation to produce more accurate predictions with minimal accumulated error. Similarly, regions corresponding to \( \mathcal{M}_{{lmk}} \), which are associated with facial landmarks, exhibit more fine-grained motion and thus benefit from interpolation-based prediction. In contrast, background regions undergo relatively minor changes and are effectively handled using feature reuse alone. More related details and experiments will be discussed in Sec.~\ref{sec:tic_g}.
}

\section{Experiments}

\begin{table*}[t!]
\centering
\caption{\textcolor{black}{\textbf{Quantitative comparisons with SOTA baselines.} We evaluate our frameworkfor both self and cross reenactments on $256 \times 256$ test images. Numbers in blue and red indicate the second-best and the best results, respectively.}}
\label{tab:quant_rec}
\scalebox{0.7}
{\begin{tabular}{lcccc|ccc|ccc}
\toprule
\multirow{2}{*}{Method}  & \multicolumn{4}{c}{\textbf{Self Reenactment}} & \multicolumn{3}{c}{\textbf{Cross Reenactment}} & \multicolumn{3}{c}{\textbf{User Study}} \\ \cmidrule(lr){2-5} \cmidrule(lr){6-8} \cmidrule(lr){9-11}
 & \textbf{L1}\ $\downarrow$  & \textbf{SSIM}\ $\uparrow$  & \textbf{LPIPS}\ $\downarrow$  & \textbf{FVD}\ $\downarrow$ & \textbf{ID Similarity}\ $\uparrow$ &\textbf{Image Quality}\ $\uparrow$ & \textbf{Landmark Accuracy}\ $\uparrow$ & \textbf{Expression}\ $\downarrow$ & \textbf{Identity}\ $\downarrow$ & \textbf{Overall}\ $\downarrow$\\
\midrule
Face Vid2vid~\cite{wang2021facevid2vid}   &0.044    & 0.823 & 0.246 & 237.7 & 0.713 & 37.294 & 5.42 & 5.42 & 6.93 &  5.87 \\
DaGAN~\cite{hong2022depth}  &0.065    & 0.743 & 0.324 & 226.4 & 0.428 & 32.168 & 1.07 &6.76 & 6.46 & 6.95  \\
TPS~\cite{TPS}   &0.042    & 0.817 & 0.217 & 205.1 & 0.561 & 35.247 & 14.28 &5.12 & 4.91 & 4.73 \\
MCNet~\cite{MCNet}   &0.036    & 0.828 & 0.206 & 213.3 & 0.437 & 36.451 & 16.93 & 4.82 & 5.01 & 5.82 \\
FADM~\cite{zeng2023face}  &0.047    & 0.725 & 0.279 & 192.7 & 0.672 & 38.188 & 1.26 &3.21 & 3.72 & 3.66 \\
MagicDance~\cite{magicpose}   &0.041    & 0.774 & 0.193 & 146.1 & 0.686 & 40.229 & 32.72 &2.02 & 2.43 & 2.84 \\
AniPortrait~\cite{wei2024aniportrait}  & 0.032 & 0.814   & 0.144  & 135.4 & 0.753   & 52.897& 34.14 & 1.71& 1.65& 2.02\\
FollowYourEmoji~\cite{ma2024followyouremoji}  & \textcolor{myblue}{\textbf{0.026}}  & \textcolor{myblue}{\textbf{0.826}} & \textcolor{myblue}{\textbf{0.138}} & \textcolor{myblue}{\textbf{118.7}} & \textcolor{myblue}{\textbf{0.758}} & \textcolor{myblue}{\textbf{59.452}} & \textcolor{myblue}{\textbf{37.18}} &\textcolor{myblue}{\textbf{1.33}} & \textcolor{myblue}{\textbf{1.58}}& \textcolor{myblue}{\textbf{1.91}}\\ 
\midrule
Ours  & \textcolor{myred}{\textbf{0.027}}    & \textcolor{myred}{\textbf{0.829}} & \textcolor{myred}{\textbf{0.137}} & \textcolor{myred}{\textbf{115.4}}  & \textcolor{myred}{\textbf{0.761}} & \textcolor{myred}{\textbf{60.147}} & \textcolor{myred}{\textbf{38.26}} &\textcolor{myred}{\textbf{1.17}} & \textcolor{myred}{\textbf{1.42}} & \textcolor{myred}{\textbf{1.67}} \\ 
\bottomrule
\end{tabular}}
\vspace{-3mm}
\end{table*}

\subsection{Implementation Details}
\textcolor{myyellow}{We jointly train our model on HDTF~\cite{zhang2021flow}, VFHQ~\cite{xie2022vfhq}, and our collected dataset comprising 18 posed expressions and 20-minute video recordings of 115 subjects captured indoors (VFHQ provides the outdoor scenes). The training pipeline unfolds in two phases. In Phase I, we randomly sample individual frames, resize and center-crop them to $512\times512$, then fine-tune for 30{,}000 iterations with a batch size of 32. Phase II focuses on temporal modeling: we train the temporal attention layers for 10{,}000 steps on 16-frame clips, also with batch size 32. The input images are sourced from Civitai~\cite{civitai,WaiRealmix,DuchaitenPonyReal}. We employ a constant learning rate of $1\times10^{-5}$ throughout both phases. The temporal attention weights are initialized from AnimateDiff~\cite{guo2023animatediff}, analogous to AnimeAnyone, while the Stable Diffusion autoencoder remains frozen to encode each frame independently. Optimization uses Adam~\cite{loshchilov2017decoupled} across 32 NVIDIA A800 GPUs over approximately 68 hours. At inference time, we utilize a DDIM sampler~\cite{song2020denoising} with classifier-free guidance scale set to 3.5.}

\textcolor{black}{
}

\subsection{EmojiBench++}
 \begin{figure}[ht]
  \centering
  \includegraphics[width=1.0\linewidth]{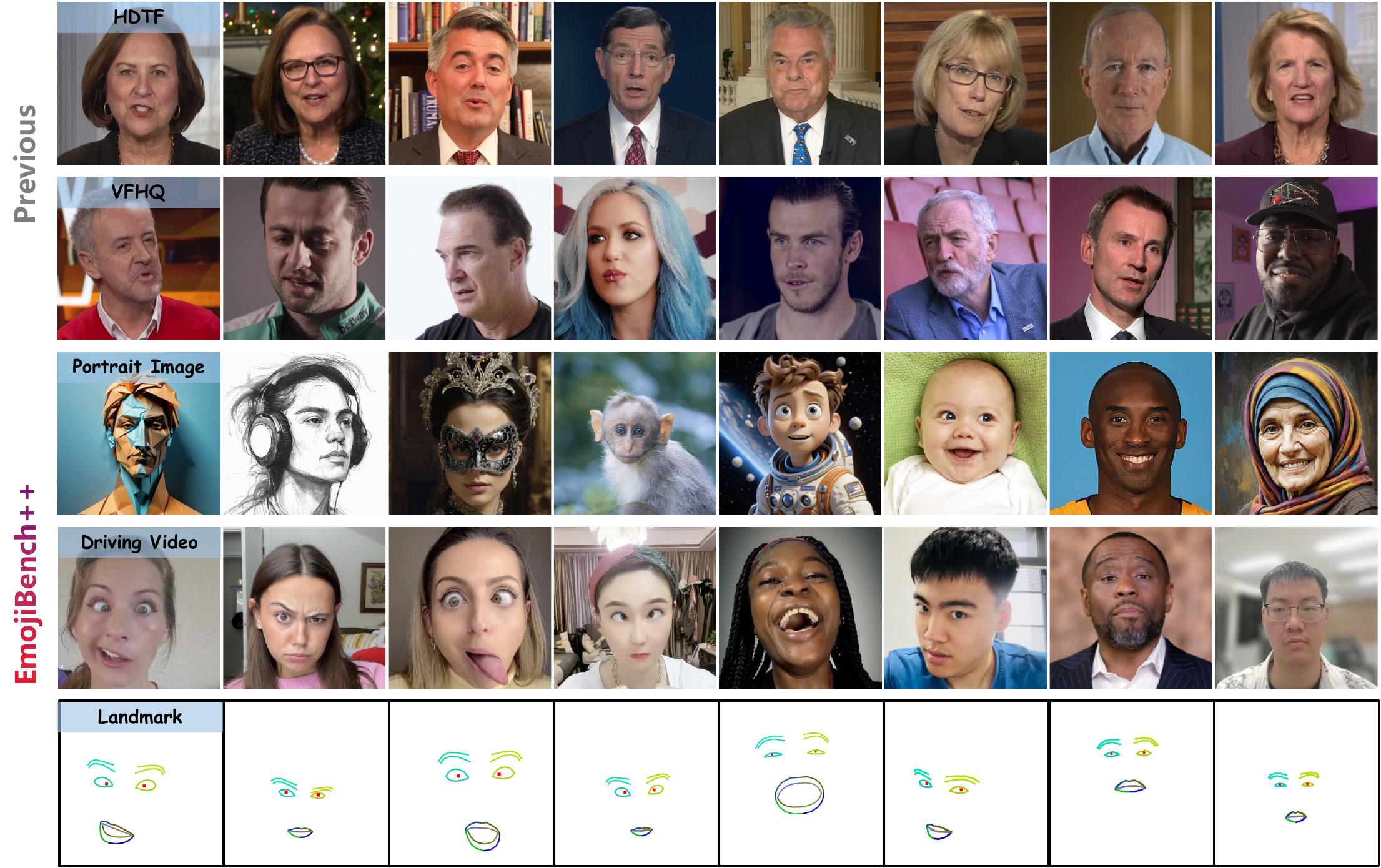} 
  \caption{\textbf{Examples of the EmojiBench++}. We collected \textcolor{myyellow}{500} portraits with high expression diversity, exaggeration, and various visual styles.
  }
  \vspace{-0.18in}
  \label{fig:emojibench}
\end{figure}

\begin{figure}
    \centering
    \includegraphics[width=1\linewidth]{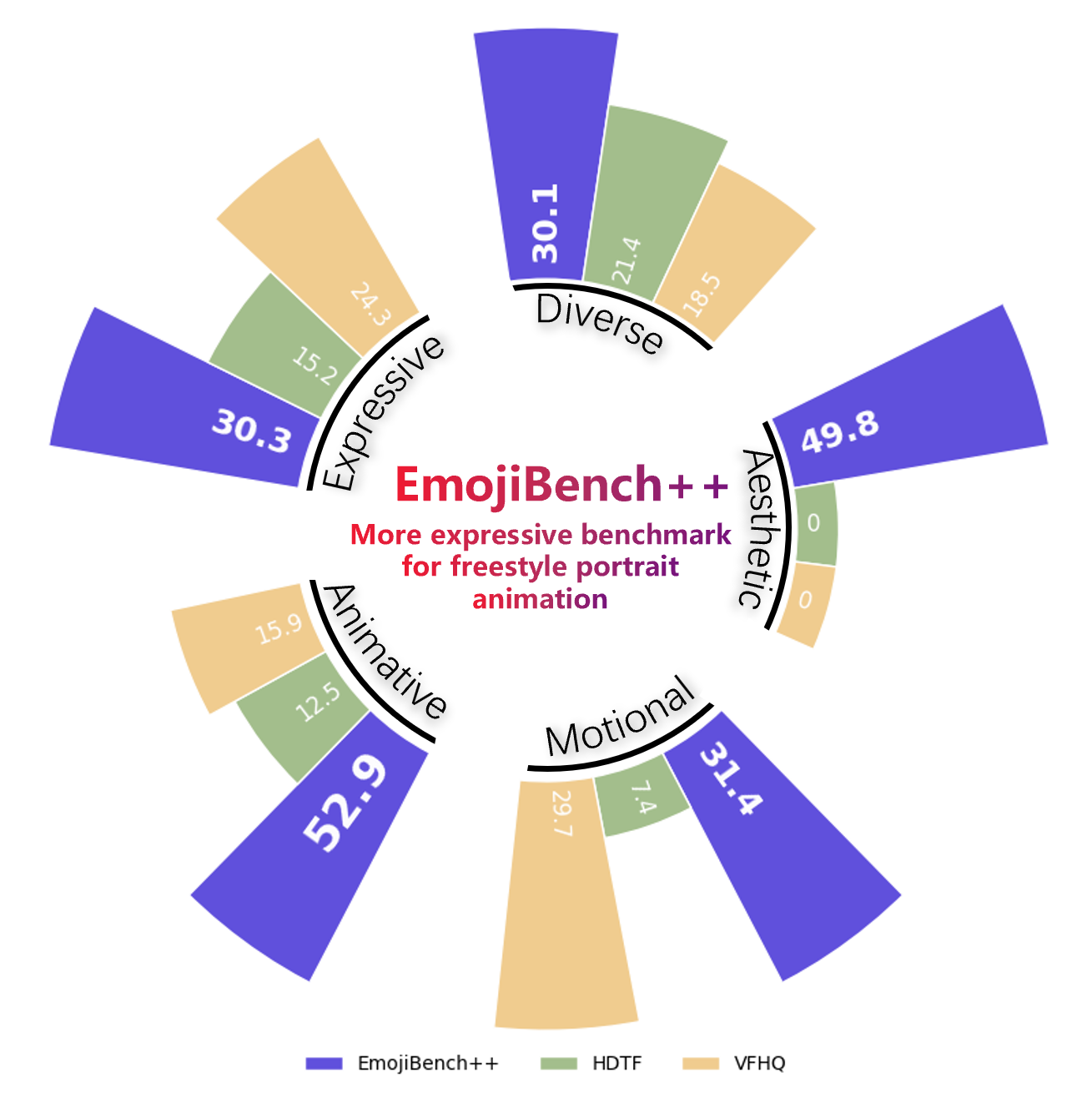}
    \caption{Comparative score visualization for EmojiBench++.}
    \label{fig:emojibenchstat}
\end{figure}

\begin{table*}[t]
\caption{Comparison of different animation datasets. }
\centering
\resizebox{\textwidth}{!}{
    \begin{tabular}{m{0.28\linewidth} m{0.1\linewidth} m{0.1\linewidth} m{0.1\linewidth} m{0.03\linewidth} m{0.15\linewidth} m{0.25\linewidth} m{0.15\linewidth}}
    \toprule
    \multicolumn{1}{c}{\multirow{2}{*}{\textbf{Datasets}}} & \multicolumn{1}{c}{\multirow{2}{*}{\textbf{Talking Head?}}} & \multirow{2}{*}{\textbf{\begin{tabular}[c]{@{}c@{}}Landmark?\end{tabular}}} & \multicolumn{1}{c}{\multirow{2}{*}{\textbf{Anime Style?}}} &\multicolumn{1}{c}{\multirow{2}{*}{\textbf{Year}}} & \multicolumn{3}{c}{\textbf{Diversity Quality}}                                \\

    \multicolumn{1}{c}{}                                   &     &                           &                                                   &                                    & \multicolumn{1}{c}{\textbf{Resolution}} & \multicolumn{1}{c}{\textbf{Nationality}}                                           & \multicolumn{1}{c}{\textbf{Expression}} \\ \midrule
    
    HDTF\cite{zhang2021flow}   & \multicolumn{1}{c}{\cmark }   & \multicolumn{1}{c}{\xmark} & \multicolumn{1}{c}{\xmark}         & \multicolumn{1}{c}{2021}                             &  \multicolumn{1}{c}{\rev{3}}   & \multicolumn{1}{c}{\rev{2.5}}           &   \multicolumn{1}{c}{\rev{1.5}}   \\

    VFHQ\cite{xie2022vfhq}       & \multicolumn{1}{c}{\cmark}   & \multicolumn{1}{c}{\xmark}  & \multicolumn{1}{c}{\xmark}  & \multicolumn{1}{c}{2022}                               &   \multicolumn{1}{c}{\rev{1.5}}   & \multicolumn{1}{c}{\rev{3}}           &   \multicolumn{1}{c}{\rev{2}} \\

    \textsc{EmojiBench++}                                 & \multicolumn{1}{c}{\cmark}   & \multicolumn{1}{c}{\cmark}         & \multicolumn{1}{c}{\cmark}       & \multicolumn{1}{c}{2025}  &    \multicolumn{1}{c}{\rev{3}}   & \multicolumn{1}{c}{\rev{3}}           &   \multicolumn{1}{c}{\rev{3}}   \\ \bottomrule
    
    \end{tabular}
}
\label{table:datasets_compare}
\end{table*}

\textcolor{myyellow}{
We introduce a more expressive and diverse free style portrait animation benchmark, EmojiBench++ as shown in Fig.~\ref{fig:emojibench}, which is expanded on original benchmark introduced in Follow-Your-Emoji~\cite{Followyouremoji}. 
The extended Benchmark++ encompasses the entirety of the original dataset and preserves its favorable characteristics. It comprises a larger collection of portrait images drawn from diverse domains, including cartoon renderings, real‐world human faces, and animal heads. These portraits originate from various personalized text‐to‐image generation models, publicly available online repositories, and user‐contributed uploads; all images have been processed with MediaPipe to extract facial landmarks~\cite{mediapipe}. 
EmojiBench++ places greater emphasis on both diversity and comprehensiveness, exhibiting a wider range of head motions and facial expressions than its predecessor. It should be noted that,  EmojiBench++ is intended solely for evaluation purposes and does not overlap with any training datasets. In summary, EmojiBench++ fully subsumes the content and distribution of EmojiBench while offering a richer, more varied content.
}
\begin{figure*}[t]
  \centering
  \includegraphics[width=1.0\linewidth]{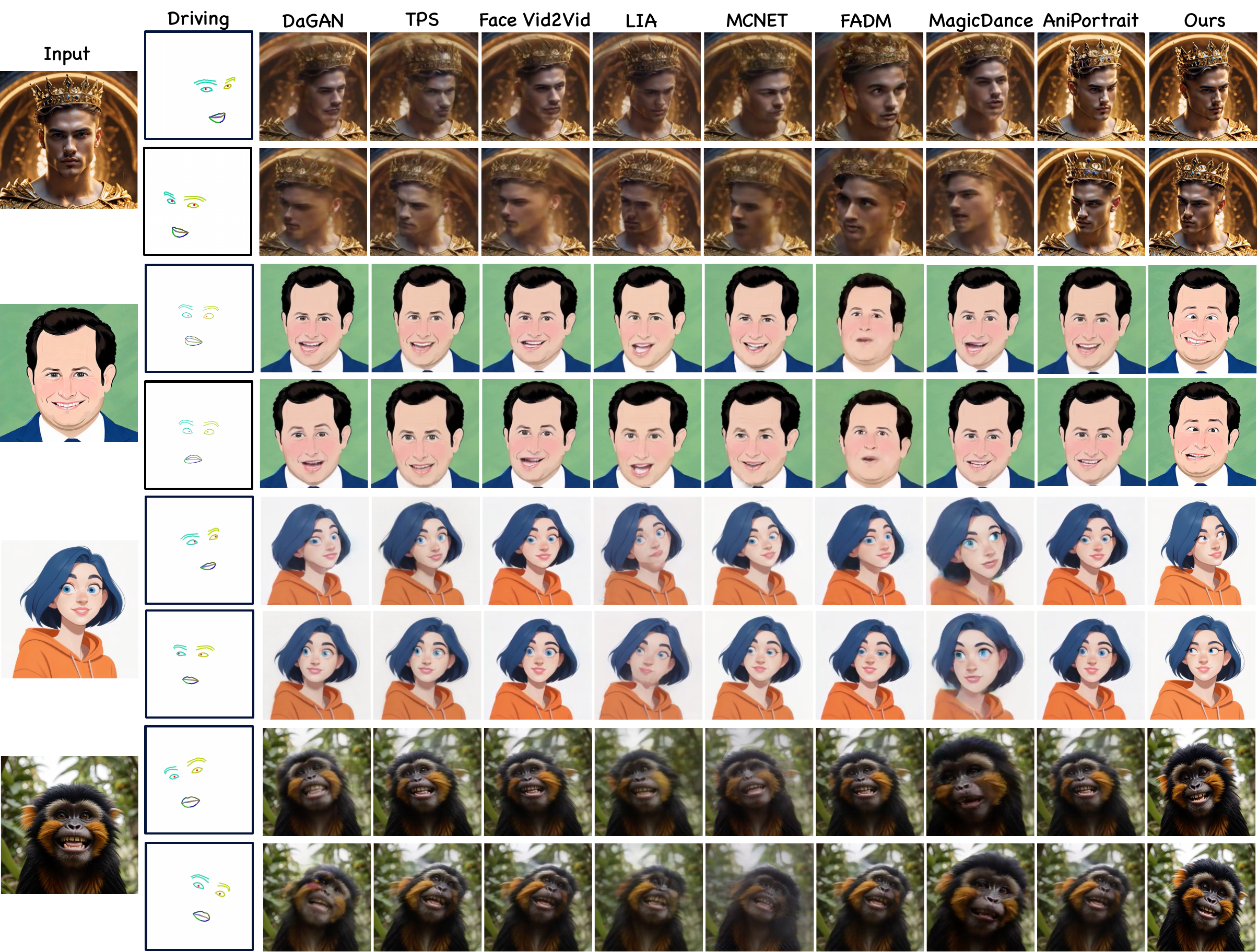} 
  \caption{\textbf{The qualitative comparisons with existing methods.} Given a reference portrait image and expression-aware landmarks, our approach demonstrates superior performance in capturing detailed facial expressions and maintaining the original identity of the characters compared to previous methods. More results are available in the supplementary material. The input images are from Civitai~\cite{civitai}.}

  \label{fig:compare}
\end{figure*}

\begin{figure}[t]
\centering
\includegraphics[width=1.0\linewidth]{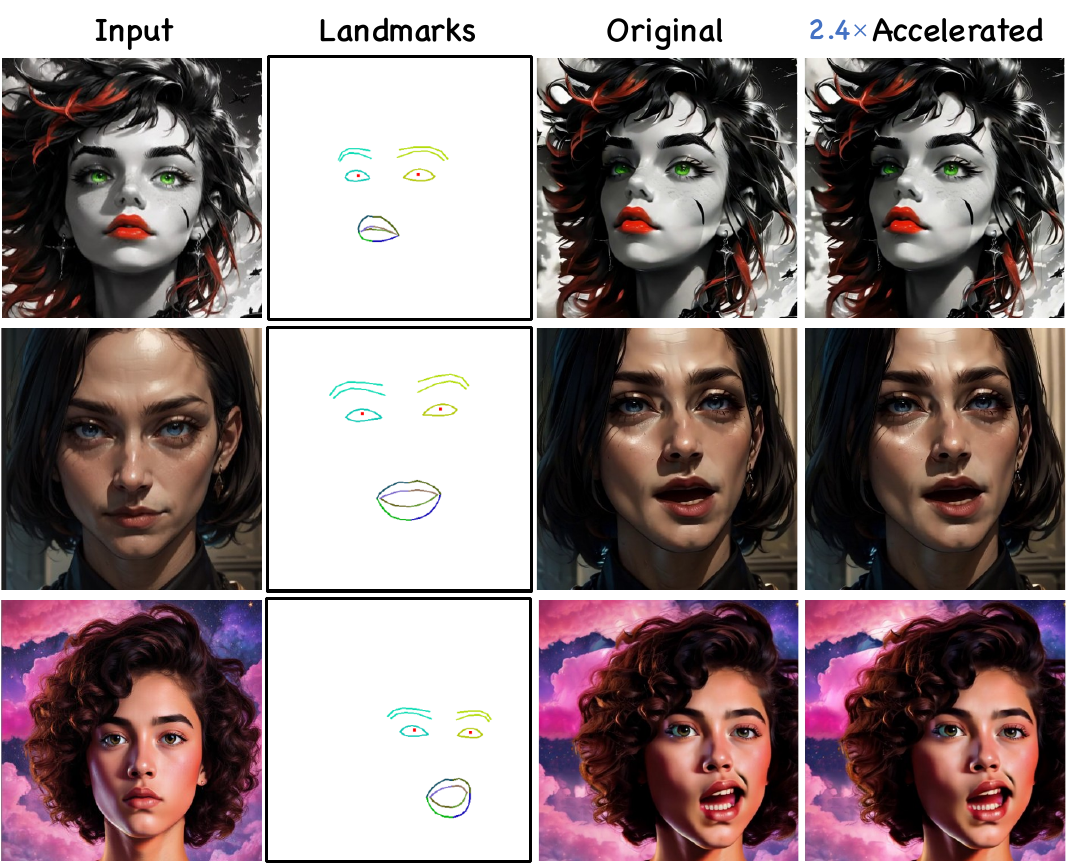} 
\caption{\textbf{Visual comparison between original output and accelerated ones.} With proposed acceleration strategy, we observe that there is almost no degradation in visual effects.}
\label{fig:accelerated_quality}
\end{figure}

\subsection{Comparison with baselines}
\label{sec:tic_g}
\subsubsection{Qualitative results}
\noindent\textbf{Comparison of Benchmarks} \textcolor{myyellow}{
We compared EmojiBench++ against two established portrait‐animation benchmarks, VFHQ~\cite{xie2022vfhq} and HQTF~\cite{zhang2021flow}, by conducting a comprehensive qualitative and visual analysis across multiple evaluation criteria. In terms of dataset composition (see Table~\ref{table:datasets_compare}), we further applied the SigLIP evaluator to score each dataset on aesthetic, dynamic, and cartoon‐style expressiveness; the resulting normalized scores are presented in Fig.~\ref{fig:emojibenchstat}. Across all measures, EmojiBench++ demonstrates markedly greater compositional diversity, higher video resolution, and superior portrait expressiveness and quality compared to the competing benchmarks.
}

\noindent\textbf{Comparison with Prior Works} \textcolor{myyellow}{
Following the same settings with Follow-Your-Emoji~\cite{Followyouremoji}, we compare our cache-accelerated approach with prior portrait animation methods visually.
We finetune all the baselines on our collected dataset.
}
\textcolor{myyellow}{
The outcomes depicted in Fig.~\ref{fig:compare} indicate that GAN‑based techniques often introduce noticeable artifacts when the head is rotated by a large angle (e.g., the first subject). Moreover, they struggle to faithfully reproduce subtle expressions in reference portraits of uncommon styles (e.g., the pupil movements in the second subject). While diffusion‑based approaches such as MagicDance~\cite{magicpose} and FADM~\cite{zeng2023face} achieve more convincing expression transfer, they still fall short in maintaining the original identity during animation. In contrast, our method demonstrates superior performance in handling large pose variations, generating fine‑grained expressions, and preserving identity even for portraits with rare styles.
}
\textcolor{black}{
Moreover, as illustrated in  Fig.~\ref{fig:accelerated_quality}, the generated results with cache-based acceleration show no perceptible differences in visual quality or detail when compared to those without acceleration, while accurately preserving consistent facial poses.
}

\subsubsection{Quantitative results}

For a quantitative comparison on EmojiBench++, we benchmark our approach against state-of-the-art portrait animation methods, including the original Follow-Your-Emoji~\cite{ma2024follow}. The results are presented in Table~\ref{tab:quant_rec}. All experiments generate 64 frames at a resolution of $256\times256$. The evaluation metrics are defined as follows:

\noindent(a) \textbf{Self reenactment}: For a quantitative evaluation of image-level quality, we employ four metrics: the L1 error, SSIM~\cite{wang2004image}, LPIPS~\cite{zhang2018unreasonable}, and FVD~\cite{unterthiner2018towards}. In the EmojiBench++ dataset, the first frame of each video serves as the reference image for generating facial expression sequences, while the following frames function as both the driving input and the ground truth.

\noindent(b) \textbf{Cross reenactment}: It describes the scenario in which the reference still and the driving landmark sequence originate from two different subjects. We assess this setting across four dimensions: identity similarity, image quality, landmark precision, and a user study.  
(1) \textit{Identity similarity}: quantified using the ArcFace score~\cite{deng2019arcface}, computed as the cosine similarity between features of the source and generated images.  
(2) \textit{Image quality}: following the protocol in~\cite{xie2024x}, we utilize HyperIQA~\cite{zhang2023liqe} to gauge visual fidelity.  
(3) \textit{Landmark precision}: treating the driving landmarks as ground truth, we detect 2D keypoints on the synthesized frames and report the mean alignment error relative to the input landmark sequences.

\noindent(c) \textbf{User study}: We conduct a user study on cross reenactment, assessing three dimensions:  
(1) \textit{Expression}: judging the fidelity of the generated facial expressions;  
(2) \textit{Identity}: measuring the similarity between each synthesized frame and the reference portrait;  
(3) \textit{Overall}: evaluating the perceived overall quality of the produced videos. Detailed procedures are described in the supplementary material.

Thirty participants were recruited for the user study, which included 45 test cases. In each case, videos produced by our method and various baselines were presented. Volunteers ranked the clips on three criteria: expression generation, identity preservation, and overall quality (lower scores denote better performance, with 1 indicating the best). We then averaged the ranks for each method. As shown in Tab.~\ref{tab:quant_rec}, our approach outperforms all baselines on seven metrics, encompassing both self- and cross-reenactment as well as the user study results.

\noindent(d) \textbf{Efficiency}: we compare latency and FLOPs in the process of denoising within kinds of accelerating methods, including naive inference with half-steps, Token Prune, Token Merge, DeepCache, and Taylor Interpolated Cache. Different configurations of the caching methods are also evaluated and visualized to analyze their impact on performance. Tab.~\ref{tab:quality_methods} demonstrates that our caching method achieves the best quality metrics among all compared methods, while incurring only minimal latency and computational overhead.
In contrast, the original pipeline with full computation achieves superior video generation quality, but suffers from the longest inference latency. Simply halving the number of inference steps significantly reduces both latency and computational cost, but at the expense of noticeable quality degradation. Moreover, other acceleration methods, while reducing latency to varying degrees, lead to a substantial drop in generation quality.

\begin{table*}[]
\caption{\textbf{Quantitative experiment on different accelerating methods} for quality retention and speed up ratio.}
\centering
  \resizebox{1\textwidth}{!}{
  \begin{tabular}{ c | c |c c c c| c c c | c  c }
    \toprule
    \multirow{2}{*}{\bf Method}& \multirow{2}{*}{\bf Steps} & \multicolumn{4}{c|}{\textbf{Self Reenactment}} & \multicolumn{3}{c|}{\textbf{Cross Reenactment}}  & \multicolumn{2}{c}{\bf Acceleration}  \\ \cmidrule(lr){3-6} \cmidrule(lr){7-9}  \cmidrule(lr){10-11}
      &  & \bf{L1$\downarrow$} & \bf{SSIM$\uparrow$} & \bf{LPIPS}\({\downarrow}\) & \textbf{FVD}\(\downarrow\) & \textbf{ID Similarity}\(\uparrow\) & \textbf{Image Quality}\(\uparrow\) & \textbf{Landmark Accuracy}\(\uparrow\) & {\bf Latency(s) $\downarrow$} & {\bf Speed $\uparrow$}\\
    \midrule
  \textbf{Original}~\cite{ma2024followyouremoji} & 30 & \textcolor{myred}{\textbf{0.026}} & \textcolor{myblue}{\textbf{0.826}} & \textcolor{myblue}{\textbf{0.138}} & \textcolor{myblue}{\textbf{118.7}} & \textcolor{myblue}{\textbf{0.758}} & \textcolor{myblue}{\textbf{59.452}} & \textcolor{myblue}{\textbf{37.18}} & 6.28 & 1.00\(\times\)  \\
  $HalfSteps$ & 15  & 0.038 & 0.746 & 0.217 &  188.6 & 0.547 & 52.187   & 28.93 & \textcolor{myblue}{\textbf{3.87}} & \textcolor{myblue}{\textbf{1.62}\(\times\)}  \\ 

  TokenPrune~\cite{zhang2025sito} &  30  & 0.036 & 0.781 & 0.185 & 175.3 &0.624 & 53.265 & 30.16 & 5.56 & 1.13\(\times\)    \\
  TokenMerge~\cite{bolya2023tokenmergingvitfaster} & 30& 0.031 & 0.805 & 0.164 & 162.8 & 0.633 & 54.924 & 32.47 & 5.24 & 1.20\(\times\)  \\
  DeepCache~\cite{deepcache} & 30  & 0.029 & 0.799 & 0.149 & 153.1 & 0.642 & 55.268 & 33.54 & 4.41 & 1.42\(\times\)  \\ \midrule
  Ours & 30 & \textcolor{myblue}{\textbf{0.027}} & \textcolor{myred}{\textbf{0.829}} & \textcolor{myred}{\textbf{0.137}} & \textcolor{myred}{\textbf{115.4}} & \textcolor{myred}{\textbf{0.761}} & \textcolor{myred}{\textbf{60.147}} & \textcolor{myred}{\textbf{38.26}} & \textcolor{myred}{\textbf{2.36}} & \textcolor{myred}{\textbf{2.66}\(\times\)}  \\
    \bottomrule
  \end{tabular}
  }
  \label{tab:quality_methods}
\end{table*}

\subsection{Ablation Study}

\begin{table*}[t!]
\centering
\caption{ \textbf{Quantitative results of ablation study.} All metrics are evaluated on $256 \times 256$ test images. ${\uparrow}$ indicates higher is better. ${\downarrow}$ indicates lower is better.}
\vspace{-1mm}
\scalebox{0.8}
{
\begin{tabular}{lcccc|ccc}
\toprule
\multirow{2}{*}{Method}  & \multicolumn{4}{c}{\textbf{Self Reenactment}} & \multicolumn{3}{c}{\textbf{Cross Reenactment}} \\ \cmidrule(lr){2-5} \cmidrule(lr){6-8}
 & \textbf{L1}\ $\downarrow$  & \textbf{SSIM}\ $\uparrow$  & \textbf{LPIPS}\ $\downarrow$  & \textbf{FVD}\ $\downarrow$ & \textbf{ID Similarity}\ $\uparrow$ &\textbf{Image Quality}\ $\uparrow$ & \textbf{Landmark Accuracy}\ $\uparrow$ \\
\midrule
FFG Loss (w/o Expression Mask)  & 0.038 & 0.710 & 0.164 & 151.3 & 0.592 & 54.814 & 27.52 \\
FFG Loss (w/o Identity Mask)  &0.037 & 0.737 & 0.159 & 157.6 & 0.557 & 51.298 & 35.37\\
w/o Progressive Strategy  &0.034 & 0.726 & 0.148 & 135.2 & 0.658 & 52.973 & 36.29 \\
\midrule
2D Landmarks  &0.041 & 0.734 & 0.172 & 157.8 & 0.579 & 51.713 & 15.92 \\
w Facial Contour points  &0.036 & 0.795 &0.157 & 137.5 & 0.634 & 56.819 & 36.82 \\
w/o Pupil points   &0.034 & 0.768 & 0.149 &122.9 & 0.669 & 58.215 & 34.28 \\
\midrule
w/o Taylor-Interpolation  & 0.029 & 0.782 & 0.144 & \textcolor{myblue}{\textbf{119.8}} & 0.692 & 58.785 & 35.27 \\
w/o Lmk-mask Guidance  & \textcolor{myblue}{\textbf{0.028}} & \textcolor{myblue}{\textbf{0.795}} & \textcolor{myblue}{\textbf{0.141}} & 121.3 & \textcolor{myblue}{\textbf{0.711}} & \textcolor{myblue}{\textbf{58.912}} & \textcolor{myblue}{\textbf{36.82}}\\
\midrule
Ours  & \textcolor{myred}{$\mathbf{0.027}$}& \textcolor{myred}{$\mathbf{0.829}$} & \textcolor{myred}{$\mathbf{0.137}$} & \textcolor{myred}{$\mathbf{115.4}$}  & \textcolor{myred}{$\mathbf{0.761}$} & \textcolor{myred}{$\mathbf{60.147}$} & \textcolor{myred}{$\mathbf{38.26}$} \\ 
\bottomrule
\end{tabular}
}
\label{tab:ablation}
\vspace{-3mm}
\end{table*}

\textcolor{myyellow}{
In the following section, we evaluate the impact of the expression-aware landmark, the fine-grained facial loss, and the TIC. A comprehensive discussion of the progressive strategy for long-term animation is available in the supplementary material.}

\definecolor{myred}{HTML}{bf0511}
\definecolor{myblue}{HTML}{1272b6}
\begin{table*}[]
\caption{\centering \textbf{Ablation Study} on different experiment settings for Taylor interpolated cache. }
\centering
  \resizebox{1\textwidth}{!}{
  \begin{tabular}{ c | c |c c c c| c c c | c  c  c  c }
    \toprule
    \multirow{2}{*}{\bf Method}& \multirow{2}{*}{\bf Settings} & \multicolumn{4}{c|}{\textbf{Self Reenactment}} & \multicolumn{3}{c|}{\textbf{Cross Reenactment}}  & \multicolumn{4}{c}{\bf Acceleration}  \\ \cmidrule(lr){3-6} \cmidrule(lr){7-9}  \cmidrule(lr){10-13}
      &  & \bf{L1$\downarrow$} & \bf{SSIM$\uparrow$} & \bf{LPIPS}\({\downarrow}\) & \textbf{FVD}\(\downarrow\) & \textbf{ID Similarity}\(\uparrow\) & \textbf{Image Quality}\(\uparrow\) & \textbf{Landmark Accuracy}\(\uparrow\) & {\bf Latency(s) $\downarrow$} & {\bf Speed $\uparrow$} & {\bf FLOPs(T) $\downarrow$}  & {\bf Speed $\uparrow$} \\
    \midrule
  $\textbf{Original}$ & 30 steps & \textcolor{myred}{\textbf{0.026}} & \textcolor{myblue}{\textbf{0.826}} & \textcolor{myblue}{\textbf{0.138}} &  \textcolor{myblue}{\textbf{118.7}} & \textcolor{myblue}{\textbf{0.758}} & 59.452   & 37.18 & 6.28 & 1.00\(\times\) & 3041.58 & 1.00\(\times\) \\ 
  \midrule

  $\textbf{\texttt{TIC(a)}}$ &  $\mathcal{N}=2$ & 0.026 & 0.811 & 0.147 & 122.5 & 0.744 & 59.113 & 36.79 & 3.67 & 1.71\(\times\) & 1789.34 & 1.70\(\times\)  \\
  $\textbf{\texttt{TIC(b)}}$ &  $\mathcal{N}=3$ & 0.028 & 0.798 & 0.145 & 127.3 & 0.739 & 59.287 & 36.43 & \textcolor{myblue}{\textbf{2.43}} & \textcolor{myblue}{\textbf{2.58}\(\times\)} & \textcolor{myred}{\textbf{1143.17}} & \textcolor{myred}{\textbf{2.66}\(\times\)}\\

  \midrule
  $\textbf{\texttt{TIC(c)}}$ & \(\mathcal{O}=1\) , $\mathcal{N}=2$  & 0.031 & 0.786 & 0.149 & 126.4 & 0.752 & 59.366 & 36.92 & 3.58 & 1.75\(\times\) & 1668.25 & 1.82\(\times\)\\
  $\textbf{\texttt{TIC(d)}}$ & \(\mathcal{O}=2\) , $\mathcal{N}=2$ & 0.029 & 0.789 & 0.146 & 125.1 & 0.754 & 59.481 & 37.45 & 3.53 & 1.78\(\times\) & 1664.23 & 1.83\(\times\)\\
  $\textbf{\texttt{TIC(e)}}$ & \(\mathcal{O}=3\) , $\mathcal{N}=2$  & 0.029 & 0.793 & 0.144 & 123.6 & 0.757 & \textcolor{myblue}{\textbf{59.583}} & \textcolor{myblue}{\textbf{37.83}} & 3.51 & 1.79\(\times\) & 1661.49 & 1.83\(\times\)\\
  \midrule
 
  $\textbf{\texttt{TIC(f)}}$ & $\mathcal{N}=3$ + lmk-mask  & \textcolor{myblue}{\textbf{0.027}} &\textcolor{myred}{\textbf{0.829}} & \textcolor{myred}{\textbf{0.137}} & \textcolor{myred}{\textbf{115.4}} & \textcolor{myred}{\textbf{0.761}} & \textcolor{myred}{\textbf{60.147}} & \textcolor{myred}{\textbf{38.26}} & \textcolor{myred}{\textbf{2.36}} & \textcolor{myred}{\textbf{2.66}\(\times\)} & \textcolor{myblue}{\textbf{1145.62}} & \textcolor{myblue}{\textbf{2.65}\(\times\)}\\
    \bottomrule
  \end{tabular}}
  \label{tab:ablation_TIC_metrics}
\end{table*}

\begin{table}[ht]
  \centering
 \caption{\textcolor{myyellow}{Comparison of keypoint detectors on quality metrics.}}
 \scalebox{0.9}{
 
\renewcommand{\arraystretch}{2}
    \begin{minipage}{\columnwidth}
    \centering
  \begin{tabular}{@{}c@{}cccc@{}}
    \toprule
    \textbf{Method}  & \textbf{FID}$\downarrow$ & \textbf{FVD} $\downarrow$ & \textbf{SSIM}$\uparrow$ & \textbf{Sync}$\uparrow$   \\
    \midrule

\multirow{1}{*}{\centering \shortstack{X-pose~\cite{xpose}}} & \textcolor{myred}{$\mathbf{0.19}$} & \textcolor{myred}{$\mathbf{102.7}$} & \textcolor{myred}{$\mathbf{0.836}$} & \textcolor{myred}{$\mathbf{4.934}$} \\
\midrule
\multirow{1}{*}{\centering \shortstack{FaceAlignment~\cite{bulat2017far}}} & $0.33$ & $198.2$ & $0.711$ & $4.381$\\
\midrule
\multirow{1}{*}{\centering \shortstack{MediaPipe~\cite{mediapipe}}} &\textcolor{myblue}{$\mathbf{0.27}$} & \textcolor{myblue}{$\mathbf{115.4}$} & \textcolor{myblue}{$\mathbf{0.814}$} & \textcolor{myblue}{$\mathbf{4.672}$} \\

    \bottomrule 
  \end{tabular}
\end{minipage}
 }
  \label{tab:detector}
\end{table}

\begin{figure}[t]
  \centering
  \includegraphics[width=1.0\linewidth]{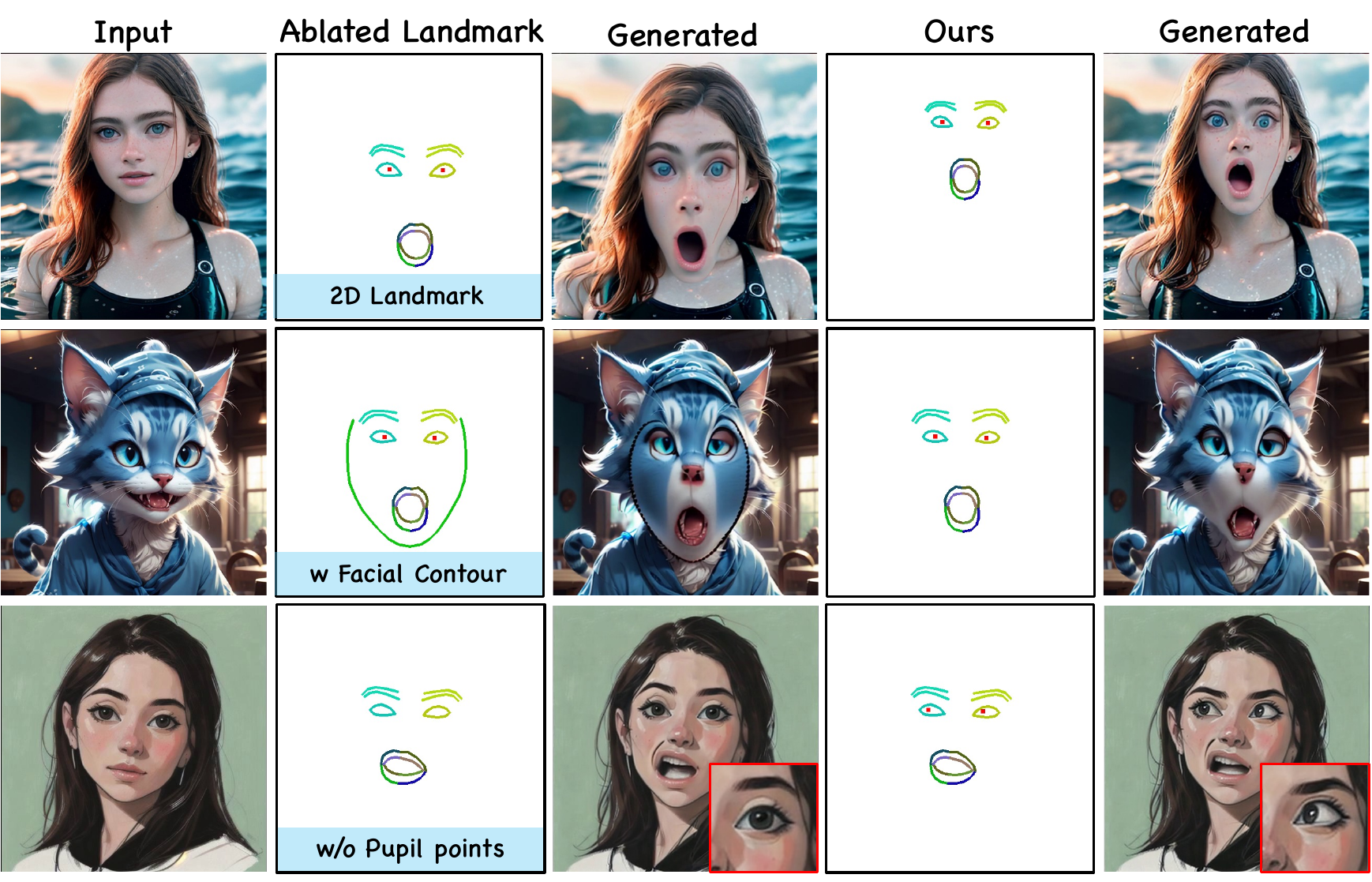} 
    \caption{\textcolor{black}{\textbf{The effectiveness of expression-aware landmark.} We compare the results when different landmarks is used to guide the portrait animation. }}
  \label{fig:ablation_lmk}
\end{figure}

\noindent\textbf{Effectiveness of expression-aware landmark.}
\textcolor{myyellow}{
 To validate the impact of our expression-aware landmarks, we substitute our motion encoding with: (i) standard 2D landmarks, (ii) expression-aware landmarks including facial contours, and (iii) expression-aware landmarks without pupil points. The visual comparisons are presented in Fig.~\ref{fig:ablation_lmk}. In the first row, plain 2D landmarks fail to align the facial bounding box between the driving landmarks and the reference portrait. When facial contours are included, identity preservation breaks down on non-human styles, owing to the inability of existing open-source detectors to accurately trace arbitrary contour shapes. Omitting pupil points removes essential motion cues, resulting in less vivid expressions. By contrast, in Tab.~\ref{tab:ablation}, our full model maintains both alignment and expressiveness.}

\begin{figure}[t]
\centering
\includegraphics[width=1.0\linewidth]{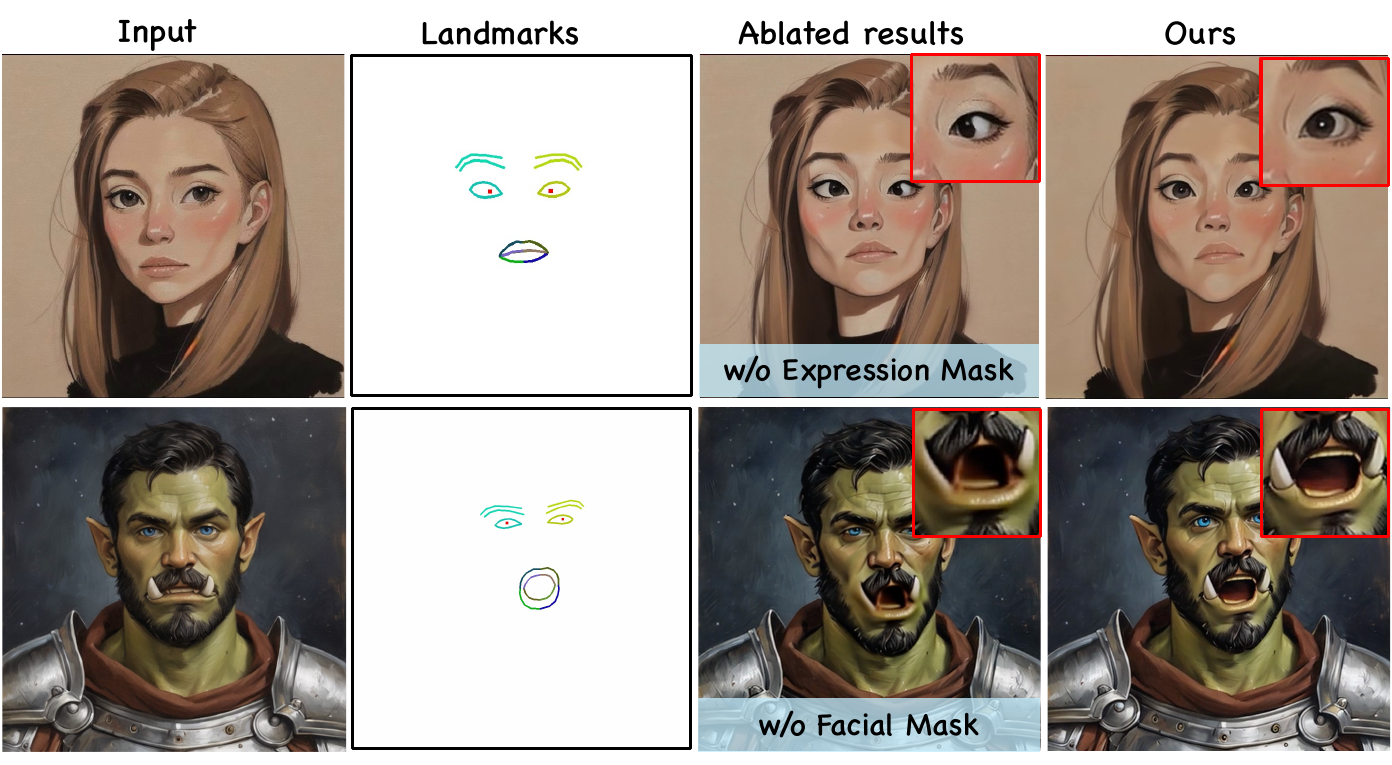} 
\caption{\textbf{The effectiveness of facial fine-grained loss.} We analyze the performance of expression and facial aspects of FFG loss, respectively. }
\label{fig:ablation_ffgloss}
\end{figure}

\noindent\textbf{Effectiveness of facial fine-grained loss.} \textcolor{myyellow}{To assess the contribution of the FFG loss, we perform ablations by removing its facial and expression components separately. Excluding the facial component compromises identity preservation and fine-grained appearance (e.g., missing teeth in the second row of Fig.~\ref{fig:ablation_ffgloss}). Omitting the expression component impairs the capture of subtle expression dynamics (e.g., inaccurate pupil motion in the first row of Fig.~\ref{fig:ablation_ffgloss}). Numerical results are summarized in Table~\ref{tab:ablation}.}

\begin{figure}
  \centering
  \includegraphics[width=\linewidth]{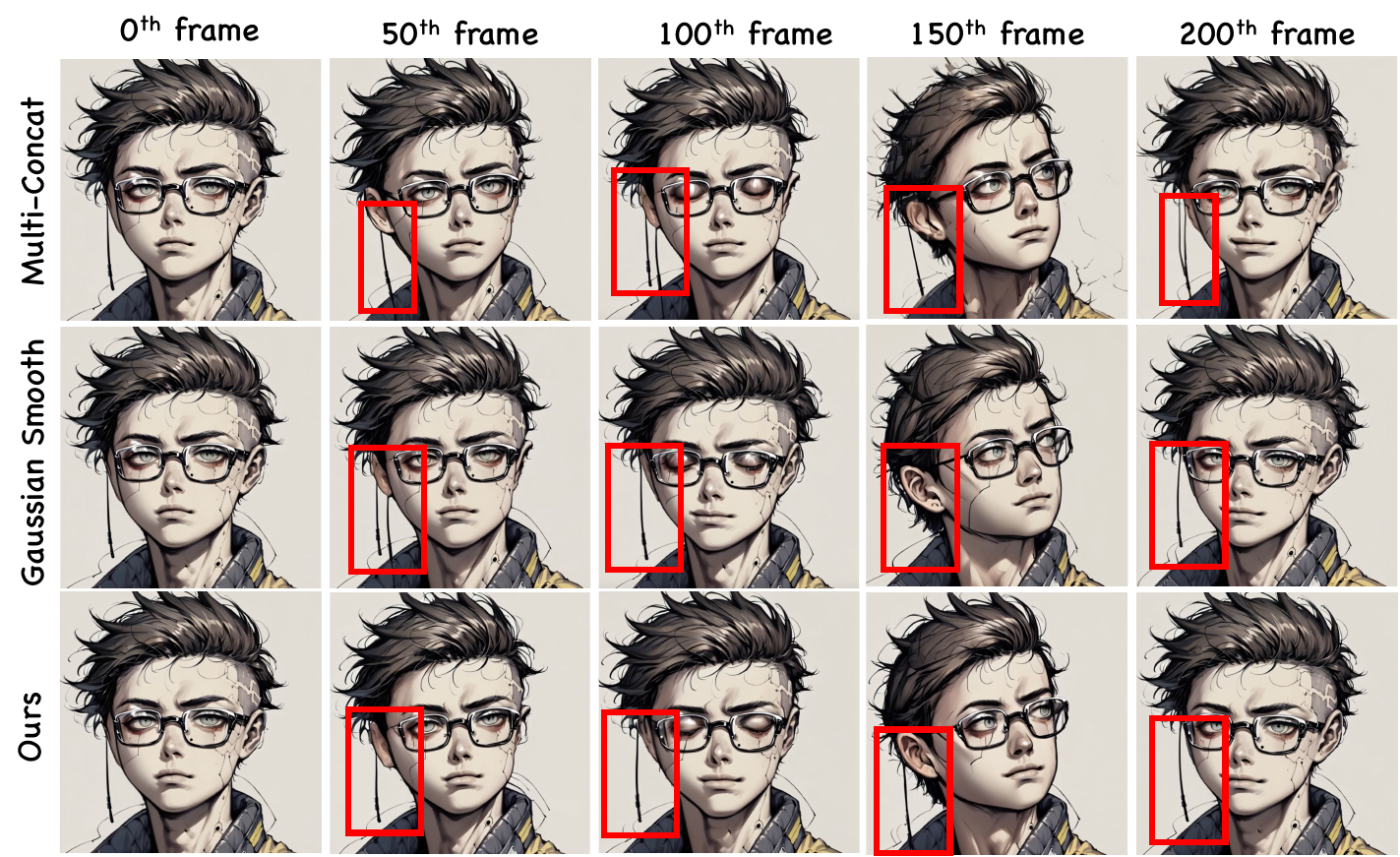} 
  \caption{The effectiveness of progressive strategy. Following the ~\cite{xu2023magicanimate}, we 
  utilize the same setting of Gaussian smoothing. As for multi-concat operation, the video clips are produced clip by clip in groups of 16 frames. 
  }
  \label{fig:ablation_progressive}
\end{figure}

\noindent\textbf{Effectiveness of progressive strategy.}\textcolor{myyellow}{We evaluate our progressive strategy for controllable long-duration video synthesis by comparing it to Gaussian smoothing and simple clip concatenation. As shown in Fig.~\ref{fig:ablation_progressive}, these alternatives fail to maintain temporal coherence and detailed identity over extended sequences. In contrast, our full method consistently preserves appearance throughout long videos, underscoring the importance of the proposed strategy.}

\begin{figure}[t]
        \centering
        \includegraphics[width=\linewidth]{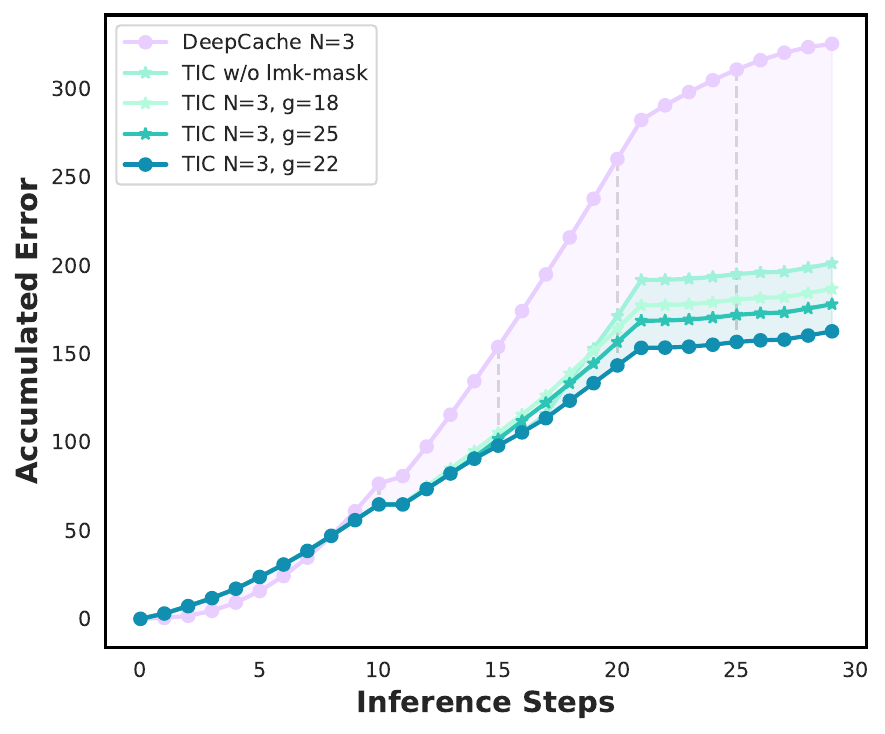}
        \caption{Ablation study showing the effect of Landmark mask and hyperparameter \(\mathcal{G}\).}
        \label{fig:ablation-cache}
\end{figure}

\begin{figure}[t]
\centering
\includegraphics[width=1.0\linewidth]{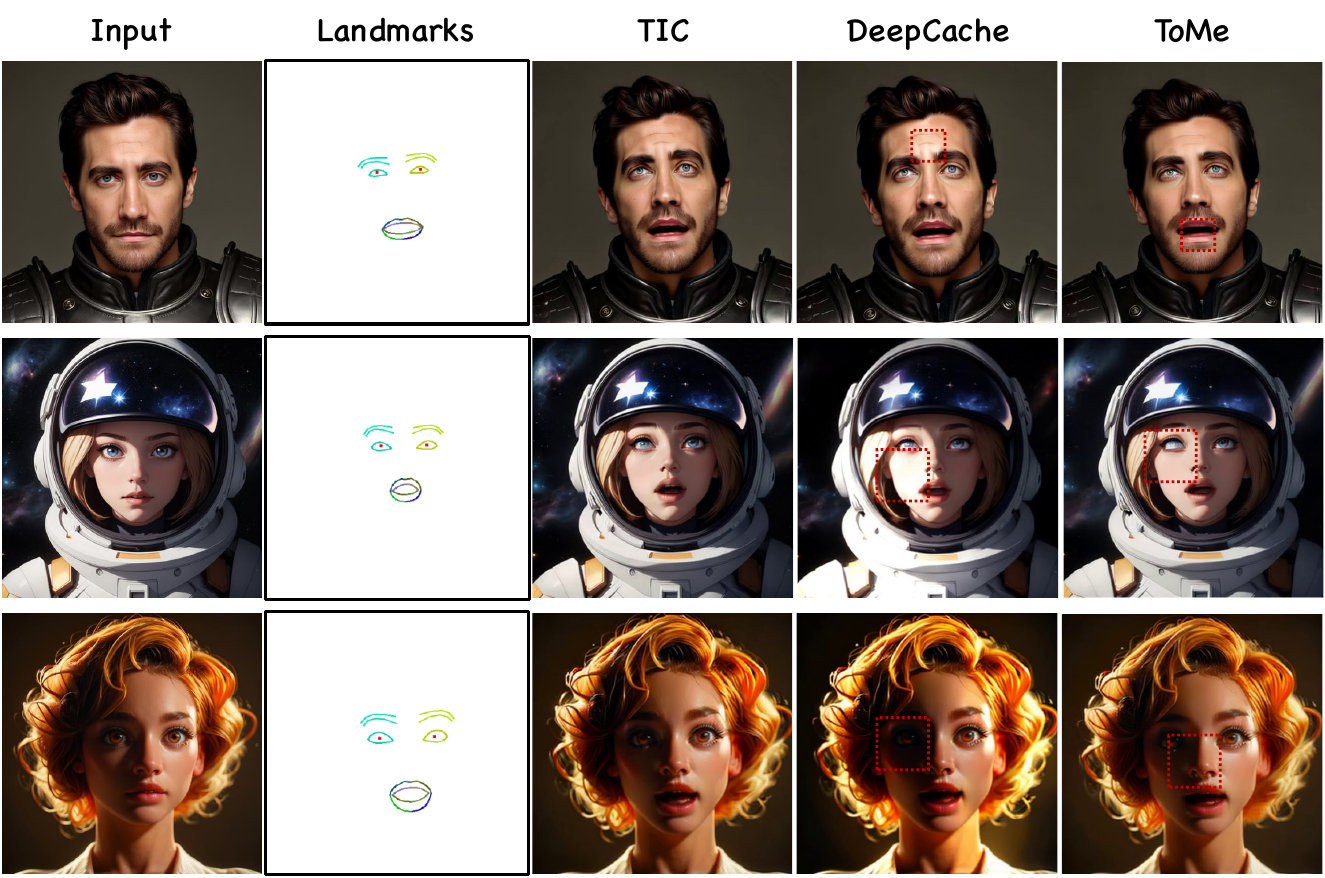} 
\caption{\textbf{Visual comparison between different acceleration methods.} Our method (TIC) demonstrates superior visual quality, while other methods exhibit varying degrees of degradation.
}
\label{fig:accelerated_ablation}
\end{figure}

\noindent\textbf{Effectiveness of Taylor Interpolated Cache.}  
\textcolor{black}{
To validate the effectiveness of the Taylor Interpolated Cache, we conduct ablation studies and compare both quality metrics and efficiency metrics under various configurations. We measure the total time overhead of the Unet denoising process as Latency. Note that we align every metric and setting with previous ones.  As shown in Tab.~\ref{tab:ablation_TIC_metrics}, increasing the refresh interval \( N \) within the cache reduces computational overhead and improves acceleration, but at the cost of slight degradation in quality metrics. 
}
\textcolor{black}{
By introducing Taylor expansion and varying its order, we are able to achieve a better trade-off between quality and speed. Finally, integrating facial landmark–based masking yields the highest acceleration ratio of 2.6× while maintaining lossless quality. These comparisons demonstrate that in the full \texttt{TIC(f)} configuration, all components work together to achieve optimal performance and efficiency, thereby validating the effectiveness of the overall design.
}

\textcolor{black}{
In addition to the ablation studies on output videos, we further investigate how different configurations affect the behavior of TIC during the generation process. We take the fully computed latents at each timestep as ground truth, and use DeepCache as a baseline under the same settings. For each configuration of TIC, we compute the mean squared error (MSE) with respect to the ground truth across timesteps.
As shown in Fig.~\ref{fig:ablation-cache}, all methods exhibit a rapid increase in cumulative error during the early timesteps, which significantly slows down after around step 20. However, the baseline shows a much faster error accumulation compared to our approach. Among the TIC variants, those without facial landmark masking accumulate more error, and different choices of the hyperparameter \( \mathcal{G} \) also lead to varying rates of error accumulation. 
The line chart indicates that there exists an optimal choice of \( \mathcal{G} \) that minimizes the rate of error accumulation, yielding results that are closer to the original, unaccelerated generation.
}

\begin{table}[t]
  \centering
 \caption{\textcolor{black}{\textbf{Quantitative results on talking-head scenarios}.}}
 \scalebox{0.9}{
 
\renewcommand{\arraystretch}{2}
    \begin{minipage}{\columnwidth}
    \centering
  \begin{tabular}{@{}c@{}cccc@{}}
    \toprule
    \textbf{Method}  & \textbf{FID}$\downarrow$ & \textbf{FVD} $\downarrow$ & \textbf{SSIM}$\uparrow$ & \textbf{Sync}$\uparrow$   \\
    \midrule

\multirow{1}{*}{\centering \shortstack{SadTalker~\cite{zhang2022sadtalker}}} & $78.291$ & $1789.236$ & $0.458$ & $1.283$ \\
\midrule
\multirow{1}{*}{\centering \shortstack{Hallo~\cite{xu2024hallo}}} & \textcolor{myblue}{$\mathbf{60.102}$} & \textcolor{myblue}{$\mathbf{1302.237}$} & $0.532$ & \textcolor{myblue}{$\mathbf{4.463}$} \\
\midrule
\multirow{1}{*}{\centering \shortstack{V-Express~\cite{wang2024V-Express}}} & $76.283$ & $2502.243$ & $0.614$ & $4.218$ \\
\midrule
\multirow{1}{*}{\centering \shortstack{AniPortrait~\cite{wei2024aniportrait}}} & $68.562$ & $2192.429$ & \textcolor{myblue}{$\mathbf{0.658}$} & $2.391$ \\
\midrule
\multirow{1}{*}{\centering \textbf{Ours}} & \textcolor{myred}{$\mathbf{56.231}$} & \textcolor{myred}{$\mathbf{967.659}$} & \textcolor{myred}{$\mathbf{0.712}$} & \textcolor{myred}{$\mathbf{4.871}$} \\

    \bottomrule 
  \end{tabular}
\end{minipage}
 }
  \label{talking-table}
\end{table}
\textcolor{black}{
Finally, we qualitatively compare the generation results of different acceleration methods. As shown in Fig.~\ref{fig:accelerated_ablation}, our proposed method (TIC) achieves the highest visual quality among all evaluated accelerate approaches. In contrast, the DeepCache-based method suffers from poor contrast, with issues such as local overexposure and excessively dark shadows, leading to significant visual degradation. The TokenMerge-based method exhibits a loss of fine details, including color blending artifacts in the eye region and blurred boundaries around the mouth and nose. These qualitative results demonstrate that, compared to other acceleration techniques, TIC not only achieves a higher speedup but also effectively overcomes the challenge of quality degradation, delivering visually satisfying results.
}

\noindent \textbf{Effectiveness of landmark detectors.}\textcolor{myyellow}{To evaluate the impact of different detectors on generation quality and to ensure the completeness of our experiments, we compare several commonly used detectors, including X-Pose~\cite{xpose}, FaceAlignment~\cite{bulat2017far}, and MediaPipe~\cite{mediapipe}. The results in Tab.~\ref{tab:detector} show that MediaPipe achieves a balanced performance among these detectors. Our workflow takes as input the facial landmarks extracted by a detector. By default, we adopt MediaPipe to extract facial skeletons, which are then used to construct valid and effective guidance inputs. However, our method does not rely on any specific detection capability and is decoupled from the performance of the detectors. In cases where invalid inputs are provided or the detector fails, the model typically produces distorted or degraded outputs.}

\section{Application}

 \begin{figure}
  \centering
  \includegraphics[width=\linewidth]{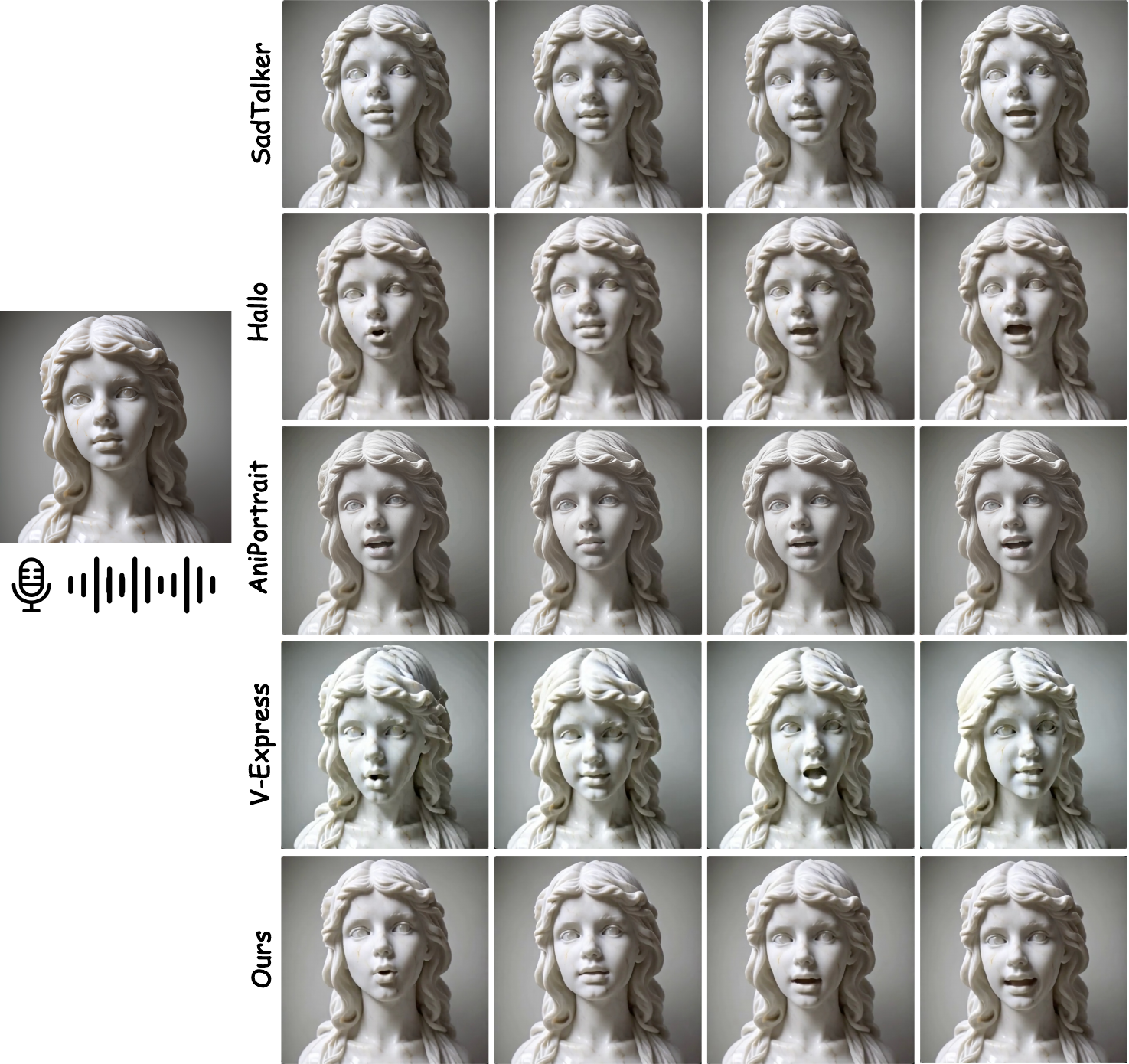} 
  \caption{\textcolor{black}{Qualitative results of the talking head scenarios. We feed the reference image and audio into the model and generate the talking head video clips. As for the landmark-guided approach, we extract the landmark from the source video using Mediapipe~\cite{mediapipe} and animate portraits.}}
  \label{fig:talking}
\end{figure}

{\noindent\textbf{Talking head.} }
We perform quantitative and qualitative comparisons on talking head scenarios. Four approaches, SadTalker\cite{zhang2022sadtalker}, Hallo~\cite{xu2024hallo}, AniPortrait~\cite{wei2024aniportrait}, V-Express~\cite{wang2024V-Express}, are selected to evaluate the ability of talking head generation. (1) Qualitative results are shown in Fig.~\ref{fig:talking}. It is easy to observe that our approach can generate consistent talking-head video. (2) Quantitative results. We evaluate these approaches using four matrixes, FID, FVD, SSIM, and Sync, respectively. As shown in Tab.~\ref{talking-table}, our approach achieves better performance on all matrices.

\begin{figure}
  \centering
  \includegraphics[width=\linewidth]{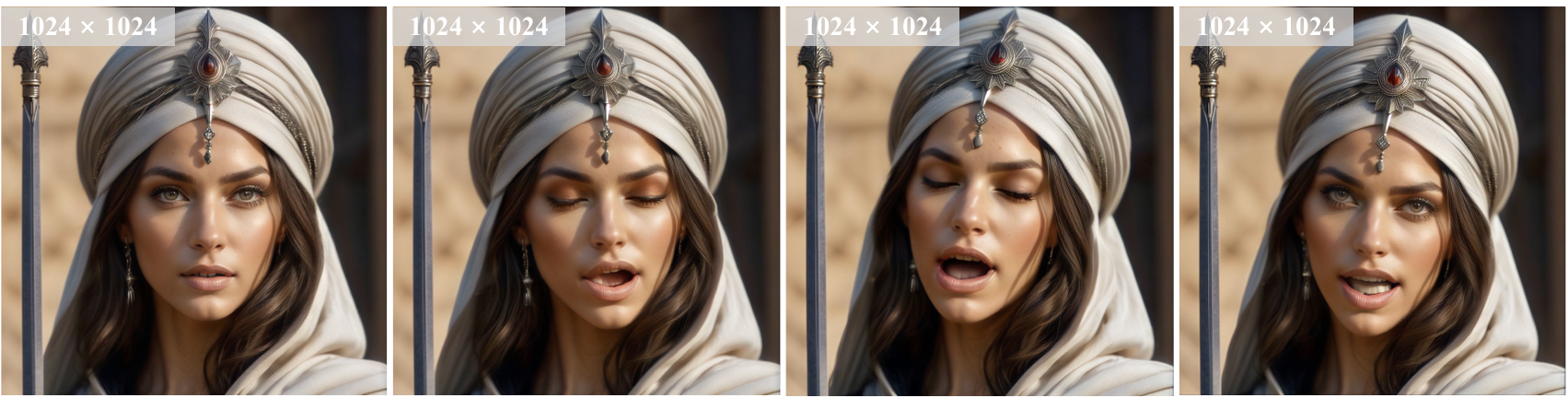} 
  \caption{\textcolor{black}{High-resolution portrait animation. Using the open-sourced approach scalecrafter~\cite{he2023scalecrafter}, we enable to animate the high-resolution portrait, such as $1024 \times 1024$. 
  }}
  \label{fig:higher}
\end{figure}

\noindent\textbf{High-resolution portrait animation.} Thanks to the current approaches in high-resolution image and video generation, we explore high-resolution animation using the proposed framework. In Fig.~\ref{fig:higher}, we provide the $1024 \times 1024$ portrait animation results. Our framework enables us to generate a consistent video clip with many details. \textcolor{myyellow}{As shown in Tab.~\ref{higher-table}, we also conducted quantitative comparisons on several high-resolution methods, including HunyuanPortrait~\cite{xu2025hunyuanportrait}, MV Portrait~\cite{lin2025mvportrait}, and WAN-2.1~\cite{wan2025}. Our approach likewise demonstrates competitive performance.}

\begin{table}[t]
  \centering
 \caption{\textcolor{myyellow}{\textbf{Quantitative results on High-resolution portrait animation}.}}
 \scalebox{0.9}{
 
\renewcommand{\arraystretch}{2}
    \begin{minipage}{\columnwidth}
    \centering
  \begin{tabular}{@{}c@{}cccc@{}}
    \toprule
    \textbf{Method}  & \textbf{FID}$\downarrow$ & \textbf{FVD} $\downarrow$ & \textbf{SSIM}$\uparrow$ & \textbf{Sync}$\uparrow$   \\
    \midrule

\multirow{1}{*}{\centering \shortstack{HunyuanPortrait
~\cite{xu2025hunyuanportrait}}} & \textcolor{myblue}{$\mathbf{76.051}$} & $850.673$ & $0.682$ & \textcolor{myblue}{$\mathbf{4.625}$} \\
\midrule
\multirow{1}{*}{\centering \shortstack{MV portrait~\cite{lin2025mvportrait}}} & $83.927$ & \textcolor{myblue}{$\mathbf{813.652}$} & $0.698$ & $4.583$ \\
\midrule
\multirow{1}{*}{\centering \shortstack{WAN-2.1~\cite{wan2025}}} & $98.103$ & $912.745$ & \textcolor{myblue}{$\mathbf{0.711}$} & $4.469$ \\
\midrule
\multirow{1}{*}{\centering \textbf{Ours}} & \textcolor{myred}{$\mathbf{67.458}$} & \textcolor{myred}{$\mathbf{720.341}$} & \textcolor{myred}{$\mathbf{0.745}$} & \textcolor{myred}{$\mathbf{5.102}$} \\

    \bottomrule 
  \end{tabular}
\end{minipage}
 }
  \label{higher-table}
\end{table}

\section{Discussion}
\begin{figure}[ht]
  \centering
  \includegraphics[width=\linewidth]{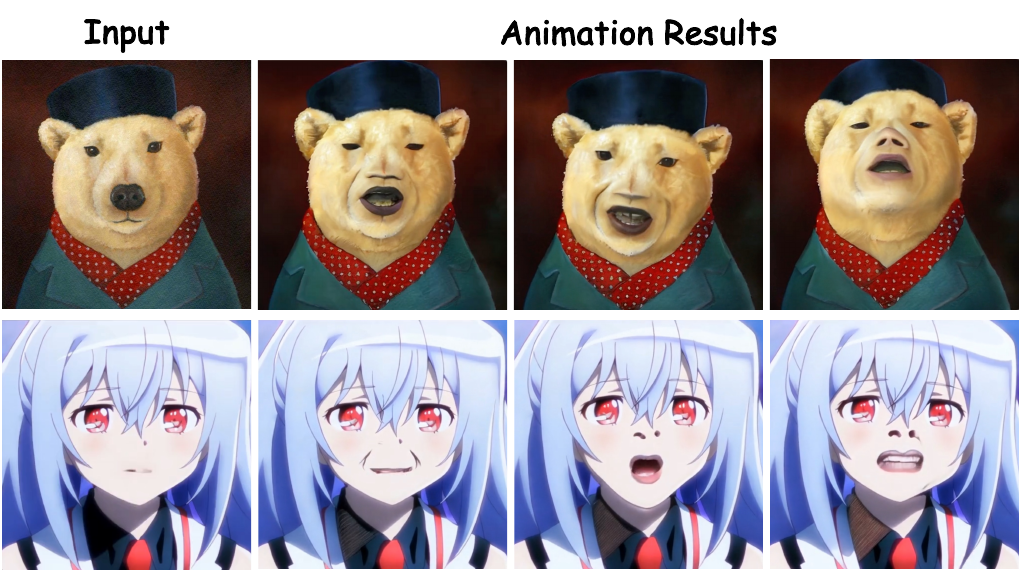} 
  \caption{\textcolor{black}{Limitation of proposed method. It is clear to observe that our approach fails to generate the details of specific domains, such as the lips of bears. Meanwhile, Follow-Your-Emoji-Faster struggles to handle portraits where landmarks cannot be detected.}}
  \label{fig:limitation}
\end{figure}
\noindent\textbf{Limitations and future work.} Although our approach enables animating freestyle portraits, it still faces the challenge of generating the vivid details of specific domain portraits, such as a cartoon bear's tongue and teeth. As shown in Fig.~\ref{fig:limitation}. This may be due to the dataset bias. The training dataset contains limited samples with these details.

\noindent\textbf{Potential negative social impact.} 
Freestyle portrait animation models offer exciting content creation opportunities but pose risks to society. The reliance on training data sourced from the internet can indeed amplify social biases, which is a concern. When machine learning models are trained on data that reflects existing societal biases, they can inadvertently perpetuate and even amplify those biases in their outputs. This is an issue that needs to be addressed to ensure fairness and equity in AI systems.

\noindent\textbf{Ethical Considerations.}
\textcolor{myyellow}{
Free-style portrait animation techniques are fundamentally challenged by the threat of deepfakes. The negative consequences of deepfakes can be far-reaching: they may mislead audiences with fabricated political news, violate individuals’ reputations, and, once disseminated broadly on social media, pose serious ethical and security risks. We expressly state that our work is intended solely for academic research and other authorized use cases, and we prohibit any malicious applications or uses that contravene applicable laws and regulations. To mitigate potential harms at a technical level, we recommend embedding a visible watermark in all generated video outputs to clearly indicate AI-generated content. In our repository, we will also prominently display a usage agreement, urging users to comply with its terms and to respect the portrait rights of others.}

\section{Conclusion} In this paper, we present Follow-Your-Emoji-Faster, an efficient diffusion-based framework for free-form portrait animation. By integrating expression-aware landmarks, our approach excels at synthesizing both subtle and exaggerated facial expressions. We also propose a facial fine-grained loss to steer the diffusion model toward preserving identity while focusing on expression generation. For training, we leverage a newly collected dataset containing 18 exaggerated expressions and 20-minute real-human videos from 115 subjects. To ensure stable long-term animations, we introduce a progressive synthesis strategy. \textcolor{myyellow}{Additionally, we construct EmojiBench++, a comprehensive benchmark with a diverse, global portrait distribution for thorough evaluation.} Experimental results confirm that our model generalizes effectively to unseen reference portraits and driving motions.
Finally, we introduce Taylor-Interpolated Cache (TIC), a plug-and-play acceleration scheme compatible with existing frameworks. TIC achieves a 2.6× speedup while maintaining generation quality without degradation.

\noindent\textbf{Acknowledgement.} This project was supported by the National Key R\&D Program of China under grant number 2022ZD0161501.

\noindent\textbf{Data Availability Statements.} All the data used in this paper can be downloaded from the Internet, and the authorization has been obtained, including civitai~\cite{civitai}, HDTF~\cite{zhang2021flow} and VFHQ~\cite{xie2022vfhq}.

{
    \small
    \bibliographystyle{ieeenat_fullname}
    \bibliography{sn-bibliography,mybib}
}


\end{document}